\documentclass[12pt]{article}

\usepackage{jmlr2e}
\usepackage{comment}
\usepackage{amsmath}
\usepackage{rotating}
\usepackage{setspace}
\usepackage{color}
\usepackage[linesnumbered, ruled]{algorithm2e}

\SetKwRepeat{Do}{do}{while}%

\newcommand{\T}{\intercal}

 \newcommand{\bma}[1]{\mbox{\boldmath $#1$}}

 \newcommand{\bX}{ {\bma{X}} }
 \newcommand{\bx}{ {\bma{x}} }

\newtheorem{assumption}[theorem]{Assumption}

\firstpageno{1}

\begin{document}

\begin{center}
  \textbf{Estimation and Optimization of Composite Outcomes} \\
	\vspace{0.1in}
  \textbf{ Daniel J.\ Luckett$^{1}$, Eric B.\ Laber$^{2}$, Michael R.\ Kosorok$^{1}$}
  \\
	\vspace{0.1in}
$^1$ Department of Biostatistics, University of North Carolina, Chapel
  Hill, NC 27599 \\ 
$^2$ Department of Statistics, North Carolina State University,
  Raleigh, NC 27695
\end{center}

\begin{abstract}
There is tremendous interest in precision medicine as a means to
improve patient outcomes by tailoring treatment to individual
characteristics.  An individualized treatment rule formalizes
precision medicine as a map from patient information to a 
recommended treatment. A treatment rule is defined to be optimal if it
maximizes the mean of a scalar outcome in a population of interest,
e.g., symptom reduction. However, clinical and intervention
scientists often must balance multiple and possibly competing
outcomes, e.g., symptom reduction and the risk of an adverse
event. One approach to precision medicine in this setting is to
elicit a composite outcome which balances all competing outcomes;
unfortunately, eliciting a composite outcome directly from patients
is difficult without a high-quality instrument, and an expert-derived
composite outcome may not account for heterogeneity in patient
preferences. We propose a new paradigm for the study of precision 
medicine using observational data that relies solely on the assumption 
that clinicians are approximately (i.e., imperfectly) making decisions to maximize
individual patient utility. Estimated composite outcomes are
subsequently used to construct an estimator of an individualized
treatment rule which maximizes the mean of patient-specific composite
outcomes. The estimated composite outcomes and
estimated optimal individualized treatment rule provide new
insights into patient preference heterogeneity, clinician behavior,
and the value of precision medicine in a given domain. We derive
inference procedures for the proposed estimators under mild
conditions and demonstrate their finite sample performance through a
suite of simulation experiments and an illustrative application to
data from a study of bipolar depression.
\end{abstract}


\section{Introduction} \label{comp.intro}

Precision medicine is an approach to healthcare that involves
tailoring treatment based on individual patient characteristics
\citep[][]{hamburg2010path, collins2015new}.  Accounting for
heterogeneity by tailoring treatment has the potential to improve
patient outcomes in many therapeutic areas. An individualized
treatment rule formalizes precision medicine as a map from the space
of patient covariates into the space of allowable treatments
\citep[][]{murphy2003optimal, robins2004optimal}.  Almost all methods
for estimating individualized treatment rules have been designed to
optimize a scalar outcome (exceptions will be discussed shortly).
However, in practice, clinical decision making often requires
balancing trade-offs between multiple outcomes. For example,
clinicians treating patients with bipolar disorder must manage both
depression and mania.  Antidepressants may help correct depressive
episodes but may also induce manic episodes
\citep[][]{sachs2007effectiveness, nassir2008antidepressants,
  goldberg2008modern, wu2015will}.  We propose a novel framework for using
observational data to estimate a composite outcome and the associated
optimal individualized treatment rule.

The estimation of optimal individualized treatment rules has been
studied extensively, leading to a wide range of estimators.  These
estimators include: regression-based methods like Q-learning 
\citep[][]{murphy2005generalization, qian2011performance, schulteOLD,
 laber2014interactive},
$A$-learning \citep[][]{murphy2003optimal, robins2004optimal, blatt2004learning, 
moodie2007demystifying, wallace2015doubly}, 
and regret regression \citep[][]{henderson2010regret};  
direct-search methods \citep[][]{rubin2012statistical, zhang2012robust, 
zhao2012estimating, zhang2013robust, zhou2015residual} based on 
inverse probability weighting 
\citep[][]{robins1999testing, murphy2001marginal, van2007causal, 
robins2008estimation}; and hybrid methods 
\citep[][]{taylor2015reader, zhang2017estimation}.     
The preceding methods require specification of a single
scalar outcome that will be used to define an optimal regime;
were individual patient utilities known, then they could be used
as the outcome in any of these methods. However, in general such 
utilities are not known though they can be elicited provided
a high-quality instrument is available
\citep[][]{butler2017incorporating}; in the absence of such an
instrument, preference elicitation is difficult to apply.  

We propose a new paradigm for estimating optimal individualized
treatment rules from observational data without eliciting patient
preferences. The key premise is that clinicians are attempting to act
optimally with respect to each patient's utility and thus the
observed treatment decisions contain information about
individual patient utilities. This idea is similar to 
that introduced by \cite{wallace2018reward} \citep[see also][]{wallace2016personalized}; 
however, we provide an estimator for the probability that a patient 
is treated optimally, rather than assuming that all patients are treated optimally.
We construct estimators of individual patient utilities which 
do not require that clinicians are acting
optimally, only that they approximately follow an optimal policy.
This approach allows us to describe the goals of the decision maker
and how these goals vary across patients, determine what makes a
patient more or less likely to be treated optimally under standard
care, and estimate the decision rule which optimizes patient-specific
composite outcomes. We develop this approach in the context of a
single-stage, binary decision in the presence of two outcomes. 
An extension to the setting with more than two outcomes is discussed in the 
Appendix. 

Other methods for estimating individualized treatment rules in the
presence of multiple outcomes include using an expert-derived
composite outcome for all patients \citep[][]{thall2002selecting,
  thall2007adaptive}.  However, this does not account for differences
in the utility function across patients and in some cases it may not
be possible to elicit a high-quality composite outcome from an
expert. Alternatively, multiple outcomes can be incorporated using
set-valued treatment regimes \citep[][]{laber2014set,
  lizotte2016multi, wu2016set}, constrained optimization
\citep[][]{linn2015chapter, laber2016identifying}, or inverse
preference elicitation \citep[][]{lizotte2012linear}.
\cite{schnell2017subgroup} extend methods for estimating the
benefiting subgroup to the case of multiple outcomes using the concept
of admissibility \citep[see also][]{schnell2016bayesian}.  However,
none of these approaches provide a method for estimating an individual
patient's utility.

This work is closely related to inverse reinforcement learning
\citep[][]{kalman1964linear, ng2000algorithms,
  abbeel2004apprenticeship, ratliff2006maximum}, which involves
studying decisions made by an expert and constructing the utility
function that is optimized by the expert's decisions.  Inverse
reinforcement learning has been successfully applied in navigation
\citep[][]{ziebart2008maximum} and human locomotion
\citep[][]{mombaur2009identifying}.  Inverse reinforcement learning
methods assume that decisions are made in a single
environment. However, in the context of precision medicine, both the
utility function and the probability of optimal treatment may vary
across patients.  Our approach is a version of inverse reinforcement
learning with multiple environments.

In Section~\ref{comp.single}, we introduce a pseudo-likelihood method to 
estimate patient utility functions from observational data. 
In Section~\ref{comp.theory}, we state a number of theoretical results 
pertaining to the proposed method, including consistency and inference for the 
maximum pseudo-likelihood estimators. 
Section~\ref{comp.simul} presents a series of simulation 
experiments used to evaluate the finite sample performance
of the proposed methods. 
Section~\ref{comp.data} presents an illustrative application 
using data from the STEP-BD bipolar disorder study.
 Conclusions and a discussion of future research 
are given in Section~\ref{comp.conclude}. Proofs are given in the appendix along 
with additional simulation results and a discussion of an extension to more 
than two outcomes.

\section{Pseudo-likelihood Estimation of Utility Functions} \label{comp.single}

Assume the available data are 
$(\bX_i, A_i, Y_i, Z_i)$, $i = 1, \ldots, n$, which comprise $n$ 
independent and identically distributed copies of 
$(\bX, A, Y, Z)$, where $\bX \in \mathcal{X} \subseteq \mathbb{R}^p$ 
are patient covariates, $A \in \mathcal{A} = \left\{-1, 1\right\}$ 
is a binary treatment, and $Y$ and $Z$ are two real-valued outcomes 
for which higher values are more desirable. 
The extension to scenarios with more than two outcomes is discussed in the Appendix.
An individualized treatment rule is a function $d: \mathcal{X} \rightarrow \mathcal{A}$ 
such that, under $d$, a patient presenting with covariates 
$\bX=\bx$ will be assigned to treatment $d(\bx)$.   
Let $Y^*(a)$ denote the 
potential outcome under treatment $a\in\mathcal{A}$, and for any regime $d$,
define $Y^*(d) = \sum_{a\in\mathcal{A}}Y^*(a)1\left\{d(\bX) = a\right\}$. An
optimal regime for the outcome $Y$, say $d_Y^{\mathrm{opt}}$, satisfies
$\mathbb{E}Y^*\left(d_{Y}^{\mathrm{opt}}\right) \ge \mathbb{E}Y^*(d)$ for any
other regime $d$.  The optimal regime for the outcome $Z$, say
$d_{Z}^{\mathrm{opt}}$, is defined analogously.   In order to identify 
these optimal regimes, and subsequently to identify the optimal regime
across the class of utility 
functions introduced below, we make the following assumptions.  
\noindent
\begin{assumption} \label{assume.consistency}
Consistency, $Y = Y^*(A)$ and $Z = Z^*(A)$.
\end{assumption}

\noindent
\begin{assumption} \label{assume.positivity}
Positivity, $\mathrm{Pr}(A = a | \bX = \bx) \ge c > 0$ for some constant $c$ and 
all pairs $(\bx, a) \in \mathcal{X} \times \mathcal{A} $.
\end{assumption}

\noindent
\begin{assumption} \label{assume.ignorability}
Ignorability, $\left\{Y^*(-1), Y^*(1)\right\} \bot A \, | \, \bX$ 
and $\left\{Z^*(-1), Z^*(1)\right\} \bot A \, | \, \bX$. 
\end{assumption}

\noindent
These assumptions are standard in causal inference
\citep[][]{robins2004optimal, hernan2010causal}.
Assumption~\ref{assume.ignorability} is not empirically verifiable in
observational studies \citep[][]{rosenbaum1983assessing,
  rosenbaum1984association}.

Define $Q_Y(\bx, a) = \mathbb{E}\left(Y | \bX = \bx, A = a\right)$. 
Then, under the preceding assumptions, it can be shown that
 $d_{Y}^{\mathrm{opt}}(\bx) = \arg\max_{a \in \mathcal{A}}
 Q_Y(\bx, a)$ \citep[][]{zhang2012robust}.  Similarly, 
it follows that $d_{Z}^{\mathrm{opt}}(\bx) =
\arg\max_{a\in\mathcal{A}}Q_{Z}(\bx, a)$ where $Q_{Z}(\bx, a) =
\mathbb{E}\left(Z|\bX=\bx, A=a\right)$.  
In general, $d^\mathrm{opt}_Y(\bx)$ need not equal $d^\mathrm{opt}_Z(\bx)$; 
therefore, if both $Y$ and $Z$ are clinically relevant, neither $d^\mathrm{opt}_Y$ 
nor $d^\mathrm{opt}_Z$ may be acceptable.    We assume that there
exists an unknown and possibly covariate-dependent utility
$U = u(Y,Z)$, where $u:\mathbb{R}^2\rightarrow\mathbb{R}$ measures
the ``goodness'' of the outcome pair $(y, z)$.  The optimal
 regime with respect to $U$, say $d_{U}^{\mathrm{opt}}$, satisfies
$\mathbb{E}U^*\left(d_U^{\mathrm{opt}}\right) = 
\mathbb{E}u\left\lbrace Y^*\left(d_{U}^{\mathrm{opt}}\right),
  Z^*\left(d_{U}^{\mathrm{opt}}\right)
\right\rbrace \ge 
\mathbb{E}u\left\lbrace Y^*(d),
  Z^*(d)
\right\rbrace = \mathbb{E}U^*(d)$ for any other regime $d$.  The
goal is to use the observed data to estimate the utility and 
subsequently $d_{U}^{\mathrm{opt}}$. Define 
$Q_{U}(\bx, a) = \mathbb{E}(U|\bX = \bx, A = a)$. For the class of utility
functions we consider below, $Q_U(\bx, a)$ is a (possibly
covariate-dependent) convex combination of $Q_Y(\bx, a)$ and 
$Q_{Z}(\bx, a)$ and is therefore identifiable under the stated causal
assumptions and furthermore $d_{U}^{\mathrm{opt}}(\bx) =
\arg\max_{a\in\mathcal{A}}Q_{U}(\bx, a)$.

We assume that clinicians act with the goal of optimizing 
each patient's utility and that their success in identifying the optimal 
treatment depends on individual patient characteristics. 
Therefore, we assume that the clinicians are approximately, i.e., imperfectly,
assigning treatment according to $d_{U}^{\mathrm{opt}}(\bx)$.  
If the clinician were always able to correctly identify the optimal treatment 
and assign $A = d^\mathrm{opt}_U(\bX)$ for each patient, there would be no 
need to estimate the optimal treatment policy \citep{wallace2016personalized}. 
Instead, we assume that the decisions of the clinician are imperfect and that 
$\mathrm{Pr}\left\{A = d^\mathrm{opt}_U(\bx) | \bX = \bx\right\} 
 = \mathrm{expit} \left(\bx^\intercal \beta \right)$ where 
$\beta$ is an unknown parameter. We implicitly assume throughout that 
$\bX$ may contain higher order terms, interactions, or basis functions 
constructed from the covariates. 

\subsection{Fixed Utility} \label{fixed.util}

We begin by assuming that the utility function is constant across patients 
and takes the form 
$u(y, z; \omega) = \omega y + (1 - \omega)z$ for some $\omega \in [0, 1]$. 
Lemma 1 of \cite{butler2017incorporating} states that, for a broad class of utility 
functions, the optimal individualized treatment rule is equivalent to the optimal 
rule for a utility function of this form. 
Define $Q_\omega(\bx, a) = \omega Q_Y(\bx, a) 
 + (1 - \omega) Q_Z(\bx, a)$ and define 
$d^\mathrm{opt}_\omega(\bx) = \mathrm{arg \, max}_{a \in \mathcal{A}} Q_\omega(\bx, a)$. 
Let $\widehat{Q}_{Y, n}$ and $\widehat{Q}_{Z, n}$ denote estimators 
of $Q_Y$ and $Q_Z$ obtained from regression models fit to 
the observed data \citep[][]{qian2011performance}. For a fixed value of $\omega$, let 
$\widehat{Q}_{\omega, n}(\bx, a) = \omega \widehat{Q}_{Y, n}(\bx, a) 
 + (1 - \omega) \widehat{Q}_{Z, n}(\bx, a)$ and subsequently let  
$\widehat{d}_{\omega, n}(\bx) = \mathrm{arg \, max}_{a \in \mathcal{A}}
 \widehat{Q}_{\omega, n}(\bx, a)$ be the plug-in estimator of $d^\mathrm{opt}_\omega(\bx)$. 
Given $\widehat{Q}_{Y, n}$ and $\widehat{Q}_{Z, n}$, $\widehat{d}_{\omega, n}(\bx)$ 
can be computed for each $\omega \in [0, 1]$. 

The joint distribution of $(\bX, A, Y, Z)$ is 
\begin{eqnarray*}
f(\bX, A, Y, Z) & = & f(Y, Z | \bX, A) f(A | \bX) f(\bX) \\
 & = & f(Y, Z | \bX, A) f(\bX) 
 \frac{\mathrm{exp}\left[\bX^\intercal \beta 1\left\{A = d^\mathrm{opt}_\omega(X)\right\}\right]}
 {1 + \mathrm{exp}\left( \bX^\intercal \beta \right)}. 
\end{eqnarray*}
Assuming that $f(Y, Z | \bX, A)$ and $f(\bX)$ do not depend on 
$\omega$ or $\beta$, the likelihood for $(\omega, \beta)$ is 
\begin{equation} \label{likelihood}
\mathcal{L}_n(\omega, \beta) \propto \prod_{i = 1}^n 
 \frac{\mathrm{exp}\left[\bX_i^\intercal \beta 1\left\{A_i = d^\mathrm{opt}_\omega(\bX_i)\right\}\right]}
 {1 + \mathrm{exp}\left(\bX_i^\intercal \beta \right)}, 
\end{equation}
which depends on the unknown function $d^\mathrm{opt}_\omega$. 
Plugging in $\widehat{d}_{\omega, n}$ for $d^\mathrm{opt}_\omega$ into (\ref{likelihood}) 
yields the pseudo-likelihood 
\begin{equation} \label{pseudo-likelihood}
\widehat{\mathcal{L}}_n(\omega, \beta) \propto \prod_{i = 1}^n 
 \frac{\mathrm{exp}\left[\bX_i^\intercal \beta 
 1\left\{A_i = \widehat{d}_{\omega, n}(\bX_i)\right\}\right]}
 {1 + \mathrm{exp}\left(\bX_i^\intercal \beta \right)}. 
\end{equation}
If we let $\widehat{\omega}_n$ and $\widehat{\beta}_n$ denote 
the maximum pseudo-likelihood estimators obtained by maximizing (\ref{pseudo-likelihood}), 
then an estimator of the utility function is 
$\widehat{u}_n(y, z) = u\left(y, z; \widehat{\omega}_n\right) 
 = \widehat{\omega}_n y + (1 - \widehat{\omega}_n) z$ 
and $\mathrm{expit}\left(\bx^\intercal \widehat{\beta}_n\right)$ 
is an estimator of the probability that a patient presenting with covariates 
$\bx$ would be treated optimally under standard care.  
An estimator of the optimal policy at $\bx$ is 
$\widehat{d}_{\widehat{\omega}_n, n} (\bx) = \mathrm{arg \, max}_{a \in \mathcal{A}} 
 \widehat{Q}_{\widehat{\omega}_n, n}(\bx, a)$. 

Because the pseudo-likelihood given in (\ref{pseudo-likelihood}) 
is non-smooth in $\omega$, standard gradient-based optimization algorithms 
cannot be used. However, for a given $\omega$, it is 
straightforward to compute the profile estimator
$\widehat{\beta}_n(\omega) = \mathrm{arg \, max}_{\beta \in \mathbb{R}^p} 
\widehat{\mathcal{L}}_n(\omega, \beta)$. We can compute the profile 
pseudo-likelihood estimator over a grid of values for $\omega$ and select 
the point on the grid yielding the largest pseudo-likelihood. 
The algorithm to construct $\left(\widehat{\omega}_n, \widehat{\beta}_n\right)$ 
is given in Algorithm~\ref{fixed.algorithm} below. 
\begin{algorithm}[h!]
Set a grid $0 = \omega_0 < \omega_1 < \ldots < \omega_K = 1$\;
\For{$m = 0, \ldots, K$}{
compute $\widehat{\beta}_{n}(\omega_n) =
\arg\max_{\beta\in\mathbb{R}^p}
\widehat{\mathcal{L}}_n(\omega_m, \beta)$ \label{step.log.reg}\;}
Select $\widehat{m}_n = \mathrm{arg \, max}_{0 \le m \le K} 
\widehat{\mathcal{L}}_n\left\{ \omega_m, \widehat{\beta}_n(\omega_m)\right\}$\;
Set $\left(\widehat{\omega}_n, \widehat{\beta}_n\right) 
 = \left\{ \omega_{\widehat{m}_n}, \widehat{\beta}_n\left( \omega_{\widehat{m}_n}\right)\right\}$\;
\caption{Pseudo-likelihood estimation of fixed utility function.}
\label{fixed.algorithm}
\end{algorithm}
Step~(\ref{step.log.reg}) can be accomplished using logistic regression. 
The theoretical properties of this estimator are discussed in Section~\ref{comp.theory}. 

\subsection{Patient-specific Utility} \label{patient.spec.util}
 
Outcome preferences can vary widely across patients in some application domains, including
schizophrenia \citep[][]{kinter2009identifying, strauss2010patients} and 
pain management \citep[][]{gan2004patient}. To accommodate this setting, 
we assume that the utility function takes the form
$u(y, z; \bx, \omega) = \omega(\bx) y + \left\{1 - \omega(\bx)\right\} z$ 
where $\omega : \mathcal{X} \rightarrow [0, 1]$ is a smooth function. 
For illustration, we let $\omega(\bx; \theta) = \mathrm{expit}\left(\bx^\intercal \theta \right)$ 
where $\theta$ is an unknown parameter. 
The situation of a misspecified model for the utility function is discussed in 
the Appendix. 
Define $Q_\theta(\bx, a) = \omega(\bx; \theta) Q_Y(\bx, a) 
 + \left\{1 - \omega(\bx; \theta)\right\} Q_Z(\bx, a)$ and define 
$d^\mathrm{opt}_\theta(\bx) = \mathrm{arg \, max}_{a \in \mathcal{A}} Q_\theta(\bx, a)$. 
Let $\widehat{Q}_{Y, n}$ and $\widehat{Q}_{Z, n}$ denote estimators 
of $Q_Y$ and $Q_Z$ obtained from regression models fit to 
the observed data. For a fixed value of $\theta$, let 
$\widehat{Q}_{\theta, n}(\bx, a) = \omega(\bx; \theta) \widehat{Q}_{Y, n}(\bx, a) 
 + \left\{1 - \omega(\bx; \theta)\right\}\cdot \widehat{Q}_{Z, n}(\bx, a)$ and subsequently let  
$\widehat{d}_{\theta, n}(\bx) = \mathrm{arg \, max}_{a \in \mathcal{A}}
 \widehat{Q}_{\theta, n}(\bx, a)$ be the plug-in estimator of $d^\mathrm{opt}_\theta(\bx)$. 
Assume that decisions are made according to the model 
$\mathrm{Pr}\left\{A = d^\mathrm{opt}_\theta(\bx) | \bX = \bx \right\} 
 =$ $\mathrm{expit}\left(\bx^\intercal \beta\right)$. 
We compute the estimators $\left(\widehat{\theta}_n, \widehat{\beta}_n\right)$ 
of $(\theta, \beta)$ by maximizing the pseudo-likelihood 
\begin{equation} \label{new-pseudo-likelihood}
\widehat{\mathcal{L}}_n(\theta, \beta) \propto \prod_{i = 1}^n 
 \frac{\mathrm{exp}\left[\bX_i^\intercal \beta 
 1\left\{A_i = \widehat{d}_{\theta, n}(\bX_i)\right\}\right]}
 {1 + \mathrm{exp}\left(\bX_i^\intercal \beta \right)}.
\end{equation}
An estimator for the utility function is $\widehat{u}_n(y, z; \bx) 
 = \omega\left(\bx; \widehat{\theta}_n\right) y 
 + \left\{1 - \omega\left(\bx; \widehat{\theta}_n\right)\right\} z$ 
and an estimator for the optimal decision function is 
$\widehat{d}_{\widehat{\theta}_n, n}$. 

As before, the pseudo-likelihood given in (\ref{new-pseudo-likelihood}) 
is non-smooth in $\theta$ and standard gradient-based optimization methods cannot be used. 
It is again straightforward to compute the profile pseudo-likelihood estimator
$\widehat{\beta}_n(\theta) = \mathrm{arg \, max}_{\beta \in \mathbb{R}^p} 
\widehat{\mathcal{L}}_n(\theta, \beta)$ for any $\theta \in \mathbb{R}^p$. 
However, because it is computationally infeasible to compute $\widehat{\beta}_n(\theta)$ 
for all $\theta$ on a grid if $\theta$ is of moderate dimension, we generate a random walk through the parameter space 
using the Metropolis algorithm as implemented in 
the \texttt{metrop} function in the R package \texttt{mcmc} \citep[][]{geyer2017mcmc} 
and compute the profile pseudo-likelihood for each $\theta$ on the random walk. 
Let $\widetilde{\mathcal{L}}_n(\theta) = \max_{\beta \in \mathbb{R}^p}
\widehat{\mathcal{L}}_n(\theta, \beta)$. We can compute $\widetilde{\mathcal{L}}_n(\theta) = 
\widehat{\mathcal{L}}_n\left\{\theta, \widehat{\beta}_n(\theta)\right\}$ by estimating 
$\widehat{\beta}_n(\theta)$ using logistic regression as described in Section~\ref{fixed.util}. 
The algorithm to construct a random walk through the parameter space is given 
in Algorithm~\ref{variable.algorithm} below.
\begin{algorithm}[h!]
Set a chain length, $B$, fix $\sigma^2 > 0$, and initialize $\theta^1$ to a starting value in $\mathbb{R}^p$\;
\For{$b = 2, \ldots, B$}{
Generate $\mathbf{e} \sim N(0, \sigma^2 I)$\;
Set $\widetilde{\theta}^{b + 1} = \theta^b + \mathbf{e}$\;
Compute $p = \min\left\{ \widetilde{L}_n\left(\widetilde{\theta}^{b + 1}\right) /  
\widetilde{L}_n\left(\widetilde{\theta}^{b}\right), 1\right\}$\;
Generate $U \sim U(0, 1)$; if $U \le p$, set $\theta^{b + 1} = \widetilde{\theta}^{b + 1}$; otherwise, 
set $\theta^{b + 1} = \theta^b$\;
}
\caption{Pseudo-likelihood estimation of patient-dependent utility function}
\label{variable.algorithm}
\end{algorithm}
After generating a chain $(\theta^1, \ldots, \theta^B)$, we select the $\theta^k$ that 
leads to the largest value of $\widetilde{\mathcal{L}}_n(\theta^k)$
as the maximum pseudo-likelihood estimator. Standard practice is to choose the 
variance of the proposal distribution, $\sigma^2$, so that the acceptance 
proportion is between 0.25 and 0.5 \citep[][]{geyer2017mcmc}. 

\section{Theoretical Results} \label{comp.theory}

Here we state a number of theoretical results 
pertaining to the proposed pseudo-likelihood estimation 
method for utility functions. We state results for a patient-specific 
utility function; the setting where the utility function is fixed is a special case. 
All proofs are deferred to the appendix.

We assume that 
$\mathrm{Pr}\left\{A = d_U^\mathrm{opt}(\bx) | \bX = \bx\right\} = \mathrm{expit}(\bx^\intercal \beta_0)$
and that the true utility function is 
$u(y, z; \bx, \theta_0) = \omega\left(\bX; \theta_0\right) y 
 + \left\{1 - \omega\left(\bX; \theta_0\right)\right\} z$, 
where $\omega(\bX; \theta)$ has bounded continuous derivative on compact sets and 
$d^\mathrm{opt}_{\theta_0}(\bX) = d^\mathrm{opt}_\theta(\bX)$ almost
surely implies
 $\theta = \theta_0$, 
i.e., the model introduced in Section~\ref{patient.spec.util} is 
well-defined and 
correctly specified with 
true parameters $\beta_0\in\mathbb{R}^{p}$ and $\theta_0\in\mathbb{R}^{d}$. 
We further assume that the estimators $\widehat{Q}_{Y, n}(\bx, a)$ and 
$\widehat{Q}_{Z, n}(\bx, a)$ are pointwise consistent for all ordered pairs $(\bx, a)$. 
Along with assumptions~\ref{assume.consistency}-\ref{assume.ignorability}, 
we implicitly assume that the densities $f(Y, Z | \bX, A)$ and $f(\bX)$ 
exist. The following result states the consistency of the 
maximum pseudo-likelihood estimators for the utility 
function and the probability of optimal treatment. 
The proof involves verifying the conditions of 
Theorem~2.12 of \cite{kosorok2008introduction}.  

\begin{theorem}[Consistency with patient-specific utility] \label{thm.patient.cons}
Let the maximum pseudo- likelihood estimators be as in Section~\ref{patient.spec.util}, 
$\left(\widehat{\theta}_n, \widehat{\beta}_n\right) = 
\mathrm{arg \, max}_{\theta \in \mathbb{R}^p, \beta \in \mathcal{B}} 
\widehat{\mathcal{L}}_n(\theta, \beta)$. Assume that $\mathcal{B}$ is a compact set 
with $\beta_0 \in \mathcal{B}$ and that $\|\mathbb{E}\bX\| < \infty$. 
Then, $\left\| \widehat{\theta}_n - \theta_0 \right\| 
\xrightarrow[]{P} 0$ and $\left\| \widehat{\beta}_n - \beta_0 \right\| 
\xrightarrow[]{P} 0$ as $n \rightarrow \infty$.
\end{theorem}

Let $V_\theta(d) = \mathbb{E}\left\{u(Y, Z; \bX, \theta) | A = d(\bX) \right\}$ 
be the mean composite outcome in a population where decisions 
are made according to $d$. The following result establishes the consistency of the value of the estimated 
optimal policy. The proof uses general theory developed by \cite{qian2011performance}. 

\begin{theorem}[Value consistency with patient-specific utility] \label{thm.patient.value}
Let $\widehat{\theta}_n$ be the maximum pseudo-likelihood estimator 
for $\theta$ and let $\widehat{d}_{\widehat{\theta}_n, n}$ be the 
associated estimated optimal policy. Then, under the given assumptions, 
$\left| V_{\theta_0}\left(\widehat{d}_{\widehat{\theta}_n, n}\right) 
 - V_{\theta_0}\left(d^\mathrm{opt}_{\theta_0}\right) \right| \xrightarrow[]{P} 0$ 
as $n \rightarrow \infty$.
\end{theorem}

Next, we derive the convergence rate and asymptotic distribution of $\left(\widehat{\theta}_n, \widehat{\beta}_n\right)$. 
Assume that $\mathcal{X}$ is a bounded subset of $\mathbb{R}^p$
and let $\| \cdot \|_\mathcal{X}$ be the sup norm over $\mathcal{X}$, 
i.e., for $f: \mathcal{X} \rightarrow \mathbb{R}$, $\| f\|_\mathcal{X} = \mathrm{sup}_{\bx \in \mathcal{X}}
|f(\bx)|$. 
Let $\dot{\omega}_\theta(\bx) = (\partial/ \partial \theta) \omega(\bx; \theta)$. 
Assume that $\big\| \|\dot{\omega}_{\theta_0}(\bx)\| \big\|_\mathcal{X} < \infty$ and 
that $\lim_{\theta \rightarrow \theta_0} \big\| \| \dot{\omega}_\theta(\bx) - \dot{\omega}_{\theta_0}(\bx)\| \big\|_\mathcal{X}
= 0$. Define $R_Y(\bx) = Q_Y(\bx, 1) - Q_Y(\bx, -1)$, $R_Z(\bx) = Q_Z(\bx, 1) - Q_Z(\bx, -1)$, 
and $R_0(\bx) = R_Y(\bx) - R_Z(\bx)$. Similarly, define 
$\widehat{R}_{Y, n}(\bx) = \widehat{Q}_{Y, n}(\bx, 1) - \widehat{Q}_{Y, n}(\bx, -1)$, 
$\widehat{R}_{Z, n}(\bx) = \widehat{Q}_{Z, n}(\bx, 1) - \widehat{Q}_{Z, n}(\bx, -1)$, 
and $\widehat{R}_{0, n}(\bx) = \widehat{R}_{Y, n}(\bx) - \widehat{R}_{Z, n}(\bx)$. 
Let $D_\theta(\bx) = \omega(\bx; \theta) R_Y(\bx) + \left\{1 - \omega(\bx; \theta)\right\} R_Z(\bx)$ 
and $\widehat{D}_{\theta, n}(\bx) = \omega(\bx; \theta) \widehat{R}_{Y, n}(\bx) 
+ \left\{1 - \omega(\bx; \theta)\right\} \widehat{R}_{Z, n}(\bx)$. 
Note that $d^\mathrm{opt}_\theta(\bx) = \mathrm{sign}\left\{ D_\theta(\bx)\right\}$ 
and $\widehat{d}_{\theta, n}(\bx) = \mathrm{sign}\left\{ \widehat{D}_{\theta, n}(\bx)\right\}$. 
Let $P_\beta(\bx) = \mathrm{expit}(\bx^\intercal \beta)$, 
$\psi_{i, A} = \left[1\left\{A_i = d^\mathrm{opt}_{\theta_0}(\bX_i)\right\} - P_{\beta_0}(\bX_i)\right] \bX_i$, 
$I_n(\beta) = \mathbb{E}_n \left[ P_{\beta}(\bX) \left\{1 - P_{\beta}(\bX) \right\} \bX \bX^\intercal\right]$, 
and $I_0 = \mathbb{E} \left[ P_{\beta_0}(\bX) \left\{1 - P_{\beta_0}(\bX) \right\} \bX \bX^\intercal\right]$. 
We use the following regularity conditions. 

\noindent
\begin{assumption} \label{assume.A}
There exist independent and identically distributed influence vectors $\psi_{1, Y},$ 
$\psi_{2, Y}, \ldots \in \mathbb{R}^{q_1}$, and 
$\psi_{1, Z}, \psi_{2, Z}, \ldots \in \mathbb{R}^{q_2}$, and vector basis functions $\phi_Y(\bx)$ and $\phi_Z(\bx)$ 
such that both
$$
\left\| \sqrt{n} \left\{\widehat{R}_{Y, n}(\bx) - R_{Y}(\bx)\right\} 
 - \phi_Y(\bx)^\intercal n^{-1/2} \sum_{i = 1}^n \psi_{i, Y} \right\|_\mathcal{X} = o_P(1)
$$
and
$$
\left\| \sqrt{n} \left\{\widehat{R}_{Z, n}(\bx) - R_{Z}(\bx)\right\} 
 - \phi_Z(\bx)^\intercal n^{-1/2} \sum_{i = 1}^n \psi_{i, Z} \right\|_\mathcal{X} = o_P(1).
$$
Let $Z_{Y, n} = n^{-1/2} \sum_{i = 1}^n \psi_{i, Y}$, $Z_{Z, n} = n^{-1/2} \sum_{i = 1}^n \psi_{i, Z}$, 
$Z_{A, n} = n^{-1/2} \sum_{i = 1}^n \psi_{i, A}$, and $q = q_1 + q_2$. 
Furthermore, assume that $\|R_Y(\bx)\|_\mathcal{X}$, $\|R_Z(\bx)\|_\mathcal{X}$, 
$\big\| \|\phi_Y(\bx)\| \big\|_\mathcal{X}$, and $\big\| \|\phi_Z(\bx)\| \big\|_\mathcal{X}$ 
are bounded by some $M < \infty$. 
Let $\Sigma_0 = \mathbb{E} \left[\left\{ \left( \psi_{1, Y}^\intercal, \psi_{1, Z}^\intercal, 
\psi_{1, A}^\intercal\right)^\intercal \right\}^{\otimes 2} \right]$ be positive definite and finite, 
where $u^{\otimes 2} = u u^\intercal$. 
\end{assumption}

\noindent
\begin{assumption} \label{assume.B}
The following conditions hold. 
\begin{enumerate}
\item The random variable $D_{\theta_0}(\bX)$ has a continuous density function $f$ in a 
neighborhood of 0 with $f_0 = f(0) \in (0, \infty)$; 
\item The conditional distribution of $\bX$ given that $|D_{\theta_0}(\bX)| \le \epsilon$ 
converges to a non-degenerate distribution as $\epsilon \downarrow 0$; 
\item There exist $\delta_1, \delta_2 > 0$ such that 
$$
\lim_{\epsilon \downarrow 0} \inf_{t \in S^d} \mathrm{Pr}\big[ |\bX^\intercal \beta_0| \ge \delta_1, 
|\left\{ R_Y(\bX) - R_Z(\bX) \right\} \dot{\omega}_{\theta_0}(\bX)^\intercal t| \ge \delta_1, 
|D_{\theta_0}(\bX)| \le \epsilon \big] \ge \delta_2, 
$$
where $S^d$ is the $d$-dimensional unit sphere. 
\end{enumerate}
\end{assumption}

\noindent
\begin{assumption} \label{assume.C}
Define, for $Z_Y \in \mathbb{R}^{q_1}$, $Z_Z \in \mathbb{R}^{q_2}$, and $U \in \mathbb{R}^d$,
\begin{multline} \label{k_0}
(Z_Y, Z_Z, U) \mapsto k_0(Z_Y, Z_Z, U) = \mathbb{E} \Big[ \bX \left\{2 P_{\beta_0}(\bX) - 1\right\} \cdot
\big| \omega(\bX; \theta_0) R_Y(\bX) \phi_Y(\bX)^\intercal Z_Y + \\
\left\{1 - \omega(\bX; \theta_0)\right\} R_Z(\bX) \phi_Z(\bX)^\intercal Z_Z + R_0(\bX) 
\dot{\omega}_{\theta_0}(\bX)^\intercal U \big| \Big| D_{\theta_0}(\bX) = 0 \Big].
\end{multline} 
Assume that $U \mapsto \beta_0^\intercal k_0(Z_Y, Z_Z, U)$ 
has a unique, finite minimum over $\mathbb{R}^d$ for all $\left(Z_Y^\intercal, Z_Z^\intercal\right)^\intercal 
\in \mathbb{R}^q$. 
\end{assumption}

\begin{remark} \label{justify.assumptions}
Assumption~\ref{assume.A} establishes a rate of convergence for the estimated 
Q-functions and is automatically satisfied if the Q-functions are estimated using 
linear or generalized linear models with or without interactions or higher order terms. 
Assumption~\ref{assume.B} is needed to ensure that there is positive probability of 
patients with $\bx$ values near the boundary between where each treatment is optimal. 
Assumption~\ref{assume.C} is standard in M-estimation. 
\end{remark}

Let $\left(\widehat{\theta}_n, \widehat{\beta}_n\right)$ be the maximum pseudo-likelihood estimators 
given in Section~\ref{patient.spec.util}. The following theorem
states the asymptotic distribution of $\left(\widehat{\theta}_n, \widehat{\beta}_n\right)$. 
\begin{theorem}[Asymptotic distribution] \label{asymptotic.distribution}
Under the given regularity conditions
\begin{equation} \label{limit}
\sqrt{n} \left( \begin{array}{c} \widehat{\theta}_n - \theta_0 \\ \widehat{\beta}_n - \beta_0 \end{array} \right)
\rightsquigarrow \left( \begin{array}{c} U \\ I_0^{-1} \left\{ Z_A - k_0(Z_Y, Z_Z, U)\right\} \end{array} \right)
\equiv \left( \begin{array}{c} U \\ B \end{array} \right),
\end{equation}
where 
$\left(Z_Y^\intercal, Z_Z^\intercal, Z_A^\intercal\right)^\intercal \sim N(0, \Sigma_0)$, 
and $U = \operatorname*{arg \, min}_{u \in \mathbb{R}^d} \beta_0^\intercal k_0(Z_Y, Z_Z, u)$.
\end{theorem}

Let $\underset{Z^*}{\overset{P}{\rightsquigarrow}}$ denote convergence in probability over $Z^*$, as defined in 
Section 2.2.3 and Chapter 10 of \cite{kosorok2008introduction}. 
Theorem~\ref{bootstrap} below establishes the validity of a parametric bootstrap procedure for 
approximating the sampling distribution of $\left(\widehat{\theta}_n, \widehat{\beta}_n\right)$. 
\begin{theorem}[Parametric bootstrap] \label{bootstrap}
Assume $\widehat{\Sigma}_n = \Sigma_0 + o_P(1)$ and $h_n = \widehat{v}_n n^{-1/6}$, 
where $\widehat{v}_n \xrightarrow[]{P} v_0 \in (0, \infty)$. Assume the regularity conditions 
given above hold. Let $Z^* \sim N(0, I^{r \times r})$, where $I^{r \times r}$ is an $r \times r$ identity matrix and $r = p + q$. 
Let $\widetilde{Z}_n = \widehat{\Sigma}_n^{1/2} Z^* 
 = \left(\widetilde{Z}_Y^\intercal, \widetilde{Z}_Z^\intercal, \widetilde{Z}_A^\intercal\right)^\intercal$. 
Let 
$$
\widetilde{T}_n(\bX, Z_Y, Z_Z) = \omega\left(\bX; \widehat{\theta}_n\right) \widehat{R}_{Y, n}(\bX) \phi_Y(\bX)^\intercal Z_Y 
 + \left\{ 1 - \omega\left(\bX; \widehat{\theta}_n\right)\right\} \widehat{R}_{Z, n}(\bX) \phi_Z(\bX)^\intercal Z_Z
$$ 
and define
\begin{multline*}
\widetilde{k}_n(Z_Y, Z_Z, U) = \mathbb{E}_n \Big[ 
 \bX \left\{2P_{\widehat{\beta}_n}(\bX) - 1\right\} \cdot \Big| \widetilde{T}_n(\bX, Z_Y, Z_Z) \\ 
 + \left\{ \widehat{R}_{Y, n}(\bX) - \widehat{R}_{Z, n}(\bX) \right\} \dot{\omega}_{\widehat{\theta}_n}(\bX)^\intercal U 
 \Big| \cdot h_n^{-1} \phi_0\left\{ \widehat{D}_{\widehat{\theta}_n, n}(\bX) / h_n\right\} \Big],
\end{multline*}
where $\phi_0$ is the standard normal density. Define 
$\widetilde{U}_n = \operatorname*{arg \, min}_{u \in \mathbb{R}^d} \widehat{\beta}_n^\intercal \widetilde{k}_n 
\left( \widetilde{Z}_Y, \widetilde{Z}_Z, u\right)$ 
and $\widetilde{B}_n = I_n\left(\widehat{\beta}_n\right)^{-1} 
\left\{\widetilde{Z}_A - \widetilde{k}_n\left(\widetilde{Z}_Y, \widetilde{Z}_Z, \widetilde{U}_n\right)\right\}$. 
Then, 
\begin{equation} \label{boot}
\left(\begin{array}{c} \widetilde{U}_n \\ \widetilde{B}_n \end{array} \right) \underset{Z^*}{\overset{P}{\rightsquigarrow}}
\left(\begin{array}{c} U \\ B \end{array} \right),
\end{equation}
where $(U^\intercal, B^\intercal)^\intercal$ is as defined in Theorem~\ref{asymptotic.distribution}. 
\end{theorem}
If we fix a large number of bootstrap replications, $B$, then
$\left(\widetilde{U}_{n, b}, \widetilde{B}_{n, b}\right), b = 1, \ldots, B$ will provide an 
approximation to the sampling distribution of the maximum pseudo-likelihood estimators. 
In Sections~\ref{comp.simul} and~\ref{comp.data}, we demonstrate the use of the bootstrap 
to test for heterogeneity of patient preferences. 

\section{Simulation Experiments}  \label{comp.simul}

\subsection{Fixed Utility Simulations}

To examine the finite sample performance of the proposed 
methods, we begin with the following simple generative model. Let $\bX = (X_1, \ldots, X_5)^\intercal$ 
be a vector of independent normal random 
variables with mean 0 and standard deviation 0.5. 
Let treatment be assigned according to 
$\mathrm{Pr}\left\{A = d^\mathrm{opt}_\omega(\bx) | \bX = \bx \right\} = \rho$, 
i.e., the probability that the clinician correctly identifies 
the optimal treatment is constant across patients. Let 
$\epsilon_Y$ and $\epsilon_Z$ be independent normal random variables 
with mean 0 and standard deviation 0.5 and let 
$Y = A\left(4X_1 - 2X_2 + 2\right) + \epsilon_Y$ and 
$Z = A\left(2X_1 - 4X_2 - 2\right) + \epsilon_Z$. 
We estimated $Q_Y$ and $Q_Z$ using linear models, 
implemented the proposed method for a variety of $n$, 
$\omega$, and $\rho$ values, and examined 
$\widehat{\omega}_n$, $\widehat{\rho}_n$, and 
$\widehat{d}_{\widehat{\omega}_n, n}$, across 500 Monte Carlo replications per scenario. 

Table~\ref{est_table1} contains mean estimates of $\omega$ and $\rho$ 
across replications along with the associated standard deviation across replications, 
and estimated error rate defined as the proportion of time the estimated optimal policy 
does not recommend the true optimal treatment.
\begin{table}[h!]
\centering
\begin{tabular}{ccc|ccc}
  \hline
$n$ & $\omega$ & $\rho$ & $\widehat{\omega}_n$ & $\widehat{\rho}_n$ & Error rate \\ 
  \hline
100 & 0.25 & 0.60 & 0.29 (0.22) & 0.62 (0.06) & 0.10 (0.11) \\ 
   &  & 0.80 & 0.23 (0.06) & 0.80 (0.04) & 0.03 (0.03) \\ 
   & 0.75 & 0.60 & 0.63 (0.24) & 0.62 (0.06) & 0.13 (0.13) \\ 
   &  & 0.80 & 0.73 (0.05) & 0.80 (0.04) & 0.03 (0.03) \\ 
   \hline
200 & 0.25 & 0.60 & 0.28 (0.18) & 0.61 (0.04) & 0.08 (0.09) \\ 
   &  & 0.80 & 0.24 (0.03) & 0.80 (0.03) & 0.02 (0.02) \\ 
   & 0.75 & 0.60 & 0.68 (0.18) & 0.61 (0.04) & 0.08 (0.10) \\ 
   &  & 0.80 & 0.74 (0.03) & 0.80 (0.03) & 0.02 (0.01) \\ 
   \hline
300 & 0.25 & 0.60 & 0.25 (0.11) & 0.61 (0.03) & 0.05 (0.06) \\ 
   &  & 0.80 & 0.24 (0.02) & 0.80 (0.02) & 0.01 (0.01) \\ 
   & 0.75 & 0.60 & 0.72 (0.13) & 0.61 (0.03) & 0.06 (0.07) \\ 
   &  & 0.80 & 0.74 (0.02) & 0.80 (0.02) & 0.01 (0.01) \\ 
   \hline
500 & 0.25 & 0.60 & 0.23 (0.08) & 0.60 (0.02) & 0.04 (0.04) \\ 
   &  & 0.80 & 0.24 (0.01) & 0.80 (0.02) & 0.01 (0.01) \\ 
   & 0.75 & 0.60 & 0.73 (0.08) & 0.60 (0.02) & 0.04 (0.04) \\ 
   &  & 0.80 & 0.75 (0.01) & 0.80 (0.02) & 0.01 (0.01) \\ 
   \hline
\end{tabular}
\caption{Estimation results for simulations where utility and probability of optimal treatment are fixed.} 
\label{est_table1}
\end{table}

The pseudo-likelihood method performs well at estimating both $\omega$ and $\rho$, 
with estimation improving with larger sample sizes as expected. 
Table~\ref{val_table1} contains estimated values of the true optimal policy, 
a policy where the utility function is estimated (the proposed method), 
policies estimated to maximize the two outcomes individually (corresponding to fixing 
$\omega = 1$ and $\omega = 0$), and 
the standard of care. The value of the standard of 
care is the mean composite outcome under the generative model. 
For each policy, the value is estimated by generating a testing sample of size 500 
with treatment assigned according to the policy and averaging 
utilities (calculated using the true $\omega$) in the testing set. The standard deviation 
across replications is included in parentheses.
\begin{table}[h!]
\centering
\resizebox{\columnwidth}{!}{
\begin{tabular}{ccc|ccccc}
  \hline
$n$ & $\omega$ & $\rho$ & Optimal & Estimated $\omega$ & $Y$ only & $Z$ only & Standard of care \\ 
  \hline
100 & 0.25 & 0.60 & 1.90 (0.07) & 1.75 (0.29) & 0.39 (0.12) & 1.77 (0.07) & 0.39 (0.23) \\ 
   &  & 0.80 & 1.90 (0.07) & 1.88 (0.07) & 0.39 (0.12) & 1.77 (0.07) & 1.14 (0.21) \\ 
   & 0.75 & 0.60 & 1.89 (0.07) & 1.69 (0.40) & 1.76 (0.08) & 0.39 (0.12) & 0.40 (0.23) \\ 
   &  & 0.80 & 1.89 (0.07) & 1.89 (0.07) & 1.76 (0.08) & 0.39 (0.12) & 1.15 (0.21) \\ 
   \hline
200 & 0.25 & 0.60 & 1.90 (0.07) & 1.80 (0.25) & 0.39 (0.11) & 1.77 (0.07) & 0.38 (0.17) \\ 
   &  & 0.80 & 1.90 (0.07) & 1.89 (0.06) & 0.39 (0.11) & 1.77 (0.07) & 1.15 (0.15) \\ 
   & 0.75 & 0.60 & 1.90 (0.07) & 1.79 (0.26) & 1.76 (0.07) & 0.38 (0.11) & 0.38 (0.17) \\ 
   &  & 0.80 & 1.90 (0.07) & 1.89 (0.06) & 1.76 (0.07) & 0.38 (0.11) & 1.16 (0.15) \\ 
   \hline
300 & 0.25 & 0.60 & 1.90 (0.07) & 1.86 (0.13) & 0.37 (0.11) & 1.76 (0.08) & 0.38 (0.13) \\ 
   &  & 0.80 & 1.90 (0.07) & 1.89 (0.07) & 0.37 (0.11) & 1.76 (0.08) & 1.14 (0.12) \\ 
   & 0.75 & 0.60 & 1.90 (0.06) & 1.84 (0.19) & 1.76 (0.08) & 0.39 (0.11) & 0.39 (0.13) \\ 
   &  & 0.80 & 1.90 (0.06) & 1.90 (0.07) & 1.76 (0.08) & 0.39 (0.11) & 1.15 (0.12) \\ 
   \hline
500 & 0.25 & 0.60 & 1.90 (0.06) & 1.88 (0.08) & 0.38 (0.11) & 1.77 (0.07) & 0.37 (0.11) \\ 
   &  & 0.80 & 1.90 (0.06) & 1.90 (0.06) & 0.38 (0.11) & 1.77 (0.07) & 1.13 (0.09) \\ 
   & 0.75 & 0.60 & 1.90 (0.07) & 1.88 (0.08) & 1.76 (0.08) & 0.39 (0.11) & 0.37 (0.10) \\ 
   &  & 0.80 & 1.90 (0.07) & 1.90 (0.07) & 1.76 (0.08) & 0.39 (0.11) & 1.13 (0.09) \\ 
   \hline
\end{tabular}
}
\caption{Value results for simulations where utility and probability of optimal treatment are fixed.} 
\label{val_table1}
\end{table}
The column labeled ``estimated $\omega$" refers to the proposed method. 
We see that the proposed method produces values which increase with $n$ and 
generally come close to the true optimal policy. In all 
settings, the proposed method offers significant improvement over the standard 
of care. The proposed method also offers improvement 
over policies to maximize each individual outcome. 

To further examine the performance of the proposed method, we 
allow the probability of optimal treatment to depend on 
patient covariates. Let $\mathrm{Pr}\left\{A 
 = d^\mathrm{opt}_\omega(\bX)\right\} 
 = \mathrm{expit}(0.5 + X_1)$. 
This corresponds to the case where $\beta = (0.5, 1, 0, \ldots, 0)^\intercal$, 
where the first element of $\beta$ is an intercept. 
Let $\bX$, $Y$, and $Z$ be generated as described above. 
In this generative model, the probability that a patient is 
treated optimally in standard care is larger for patients with 
positive values of $X_1$ and smaller for patients with negative 
values of $X_1$.  
We applied the proposed method to 500 replications 
of this generative model for various $n$ and $\omega$. 
Table~\ref{est_table2} contains mean estimates of $\omega$, 
root mean squared error (RMSE) of $\widehat{\beta}_n$, and the error rate. 
\begin{table}[h!]
\centering
\begin{tabular}{cc|ccc}
  \hline
$n$ & $\omega$ & $\widehat{\omega}_n$ & RMSE of $\widehat{\beta}_n$ & Error rate \\ 
  \hline
100 & 0.25 & 0.33 (0.24) & 1.32 (0.49) & 0.10 (0.14) \\ 
   & 0.75 & 0.70 (0.21) & 1.38 (0.45) & 0.11 (0.10) \\ 
  200 & 0.25 & 0.27 (0.14) & 0.80 (0.30) & 0.04 (0.08) \\ 
   & 0.75 & 0.73 (0.14) & 0.85 (0.28) & 0.06 (0.07) \\ 
  300 & 0.25 & 0.25 (0.08) & 0.59 (0.21) & 0.02 (0.05) \\ 
   & 0.75 & 0.75 (0.10) & 0.64 (0.21) & 0.05 (0.05) \\ 
  500 & 0.25 & 0.25 (0.03) & 0.43 (0.14) & 0.01 (0.01) \\ 
   & 0.75 & 0.76 (0.07) & 0.46 (0.15) & 0.03 (0.03) \\ 
   \hline
\end{tabular}
\caption{Estimation results for simulations where utility is fixed and probability of optimal treatment is variable.} 
\label{est_table2}
\end{table}
Estimation of the observational policy (as defined by $\beta$) improves 
with larger sample sizes. The probability that the estimated 
policy assigns the optimal treatment also increases with the sample size. 
The true value of $\omega$ does not affect estimation of $\omega$ or $\beta$. 

Table~\ref{val_table2} contains estimated values of the true optimal policy, 
a policy where the utility function is estimated (the proposed method), 
policies estimated to maximize each outcome individually, and 
the standard of care. Values are estimated from independent testing sets of size 500 as described above. 
The value under the standard of care is the mean composite outcome under the generative model. 
\begin{table}[h!]
\centering
\begin{tabular}{cc|ccccc}
  \hline
$n$ & $\omega$ & Optimal & Estimated $\omega$ & $Y$ only & $Z$ only & Standard of care \\ 
  \hline
100 & 0.25 & 1.90 (0.06) & 1.72 (0.41) & 0.40 (0.11) & 1.76 (0.07) & 0.33 (0.24) \\ 
   & 0.75 & 1.90 (0.06) & 1.76 (0.29) & 1.76 (0.07) & 0.38 (0.12) & 0.58 (0.24) \\ 
  200 & 0.25 & 1.90 (0.06) & 1.84 (0.24) & 0.38 (0.11) & 1.75 (0.08) & 0.32 (0.16) \\ 
   & 0.75 & 1.90 (0.06) & 1.84 (0.16) & 1.76 (0.07) & 0.38 (0.11) & 0.57 (0.16) \\ 
  300 & 0.25 & 1.89 (0.07) & 1.88 (0.14) & 0.38 (0.11) & 1.77 (0.07) & 0.32 (0.14) \\ 
   & 0.75 & 1.90 (0.07) & 1.87 (0.09) & 1.76 (0.07) & 0.39 (0.12) & 0.56 (0.14) \\ 
  500 & 0.25 & 1.90 (0.07) & 1.90 (0.06) & 0.38 (0.11) & 1.77 (0.07) & 0.33 (0.10) \\ 
   & 0.75 & 1.90 (0.07) & 1.89 (0.08) & 1.76 (0.07) & 0.39 (0.11) & 0.57 (0.10) \\ 
   \hline
\end{tabular}
\caption{Value results for simulations where utility is fixed and probability of optimal treatment is variable.} 
\label{val_table2}
\end{table}
The proposed method (found in the column labeled ``estimated $\omega$") 
produces values that are close to the true optimal policy 
in large samples and a significant improvement over standard of care in small to moderate samples.  
We note that value under the standard of care differs across $\omega$. When $\omega$ is 
close to 1, the composite outcome places more weight on $Y$, for which the 
magnitude of the association with $X_1$ is larger. Because patients with larger 
values of $X_1$ are more likely to be treated optimally in this generative model, 
the standard of care produces larger composite outcomes when $\omega$ is closer to 1. 
Likewise, the mean composite outcome under policies to maximize each individual outcome 
varies with the true value of $\omega$. 

\subsection{Patient-specific Utility Simulations}

Next, we examine the case where the utility function is allowed to vary 
across patients. Let $\bX$, $Y$, and $Z$ be generated as above. 
Again, assume that $\mathrm{Pr}\left\{A 
 = d^\mathrm{opt}_\theta(\bX)\right\} = \mathrm{expit}(0.5 + X_1)$, i.e., 
$\beta = (0.5, 1, 0, \ldots, 0)^\intercal$.
Consider the composite outcome $U = \omega(\bX; \theta) Y +$ $ \left\{1 - \omega(\bX, \theta)\right\} Z$, 
where $\omega(\bX; \theta) = \mathrm{expit}\left(1 -0.5 X_1\right)$, i.e., 
$\theta = (1, -0.5, 0, \ldots, 0)^\intercal$, where the first element of $\theta$ is an intercept. 
We implemented the proposed method for various $n$ and examined estimation 
of $\theta$ and $\beta$ across 500 replications. Each replication is based on a 
simulated Markov chain of length 10,000 as described in Section~\ref{patient.spec.util}. 
Results are summarized in Table~\ref{est_table3}. 
\begin{table}[h!]
\centering
\begin{tabular}{c|ccc}
  \hline
$n$ & RMSE of $\widehat{\theta}_n$ & RMSE of $\widehat{\beta}_n$ & Error rate \\ 
  \hline
100 & 2.42 (0.99) & 1.69 (0.62) & 0.15 (0.08) \\ 
  200 & 2.14 (0.91) & 1.00 (0.33) & 0.12 (0.06) \\ 
  300 & 1.94 (0.83) & 0.78 (0.25) & 0.10 (0.05) \\ 
  500 & 1.68 (0.74) & 0.53 (0.17) & 0.08 (0.04) \\ 
   \hline
\end{tabular}
\caption{Estimation results for simulations where both utility and probability of optimal treatment are variable.} 
\label{est_table3}
\end{table}
Larger sample sizes produce marginal decreases in the RMSE of $\widehat{\theta}_n$. 
The estimated policy assigns the true optimal treatment more than 
80\% of the time for all sample sizes and 
the error rate decreases as the sample size increases.
Table~\ref{val_table3} contains estimated values of the true optimal policy, 
the policy estimated using the proposed method, policies estimated to maximize 
each outcome individually, and standard of care. 
\begin{table}[h!]
\centering
\begin{tabular}{c|ccccc}
  \hline
$n$ & Optimal & Estimated $\omega$ & $Y$ only & $Z$ only & Standard of care \\ 
  \hline
100 & 1.74 (0.06) & 1.53 (0.19) & 1.59 (0.07) & 0.44 (0.11) & 0.51 (0.21) \\ 
  200 & 1.73 (0.06) & 1.61 (0.13) & 1.59 (0.07) & 0.44 (0.10) & 0.51 (0.15) \\ 
  300 & 1.74 (0.06) & 1.64 (0.12) & 1.59 (0.07) & 0.44 (0.10) & 0.50 (0.13) \\ 
  500 & 1.74 (0.06) & 1.68 (0.09) & 1.59 (0.07) & 0.43 (0.10) & 0.50 (0.09) \\ 
   \hline
\end{tabular}
\caption{Value results for simulations where both utility and probability of optimal treatment are variable.} 
\label{val_table3}
\end{table}
The proposed method produces policies that achieve significant 
improvement over the standard of care across sample sizes. 

Finally, we examine the performance of the parametric bootstrap as
described in Section~\ref{comp.theory}. Let $\bX$ be a bivariate
vector of normal random variables with mean 0, standard deviation 0.5,
and correlation zero. Let $Y$ and $Z$ be generated as above and let
$\beta = (2.5, 1, 0)^\intercal$ where the first element of $\beta$
is an intercept. Let $\theta_{(1)}$
be the vector $\theta$ with the first element removed. We are
interested in testing the null hypothesis
$H_0: \left\|\theta_{(1)}\right\| = 0$, which corresponds to a test
for heterogeneity of patient preferences. The table below
contains estimated power across 500 Monte Carlo replications under the
null hypothesis, where the true value is $\theta = (1, 0, 0)^\intercal$, 
and two alternative hypotheses: $H_1: \theta = (1, 4, 3)^\intercal$, 
and $H_2: \theta = (1, 6, 6)^\intercal$. All tests were conducted at level $\alpha = 0.05$
and based on 1000 bootstrap samples.
\begin{table}[h!]
\centering
\begin{tabular}{c|ccc}
  \hline
$n$ & Type 1 error & Power against $H_1$ & Power against $H_2$ \\ 
  \hline
100 & 0.014 & 0.190 & 0.272 \\ 
  200 & 0.020 & 0.409 & 0.572 \\ 
  300 & 0.048 & 0.497 & 0.649 \\ 
  500 & 0.066 & 0.528 & 0.682 \\ 
   \hline
\end{tabular}
\caption{Power of bootstrap test for homogeneity of utility function} 
\label{power_table4}
\end{table}
The proposed bootstrap procedure produces type I error rates near
nominal levels under the null and moderate power in large samples under
alternative hypotheses. 

\section{Case Study: The STEP-BD Standard Care Pathway}  \label{comp.data}

The Systematic Treatment Enhancement Program for Bipolar Disorder (STEP-BD) 
was a landmark study of the effects of antidepressants 
in patients with bipolar disorder \citep[][]{sachs2007effectiveness}. In addition to a randomized 
trial assessing outcomes for patients given an antidepressant or placebo, 
the STEP-BD study also included a large-scale observational study, the 
standard care pathway.  
As our method requires observational data on clinical decision making, 
we apply the proposed method to the observational data from the 
STEP-BD standard care pathway to estimate decision rules for the use of 
antidepressants in patients with bipolar disorder.  (Clearly, as
clinicians are not generally assigning treatment according to their
best clinical judgment in a randomized clinical trial, the proposed
method is not applicable to the randomized pathway of STEP-BD.)

Although bipolar disorder is characterized by alternating episodes of depression 
and mania, recurrent depression is the leading cause of impairment among 
patients with bipolar disorder \citep[][]{judd2002long}. However, the use of 
antidepressants has not become standard care in bipolar disorder due to the risk 
of antidepressants inducing manic episodes in certain patients 
\citep[][]{nassir2008antidepressants, goldberg2008modern}. Thus, the clinical 
decision in the treatment of bipolar disorder is whether to prescribe 
antidepressants to a specific patient in order to balance trade-offs between 
symptoms of depression, symptoms of mania, and other side effects of treatment. 
 
\begin{figure}[h!]
\centering
\includegraphics[scale = 0.4]{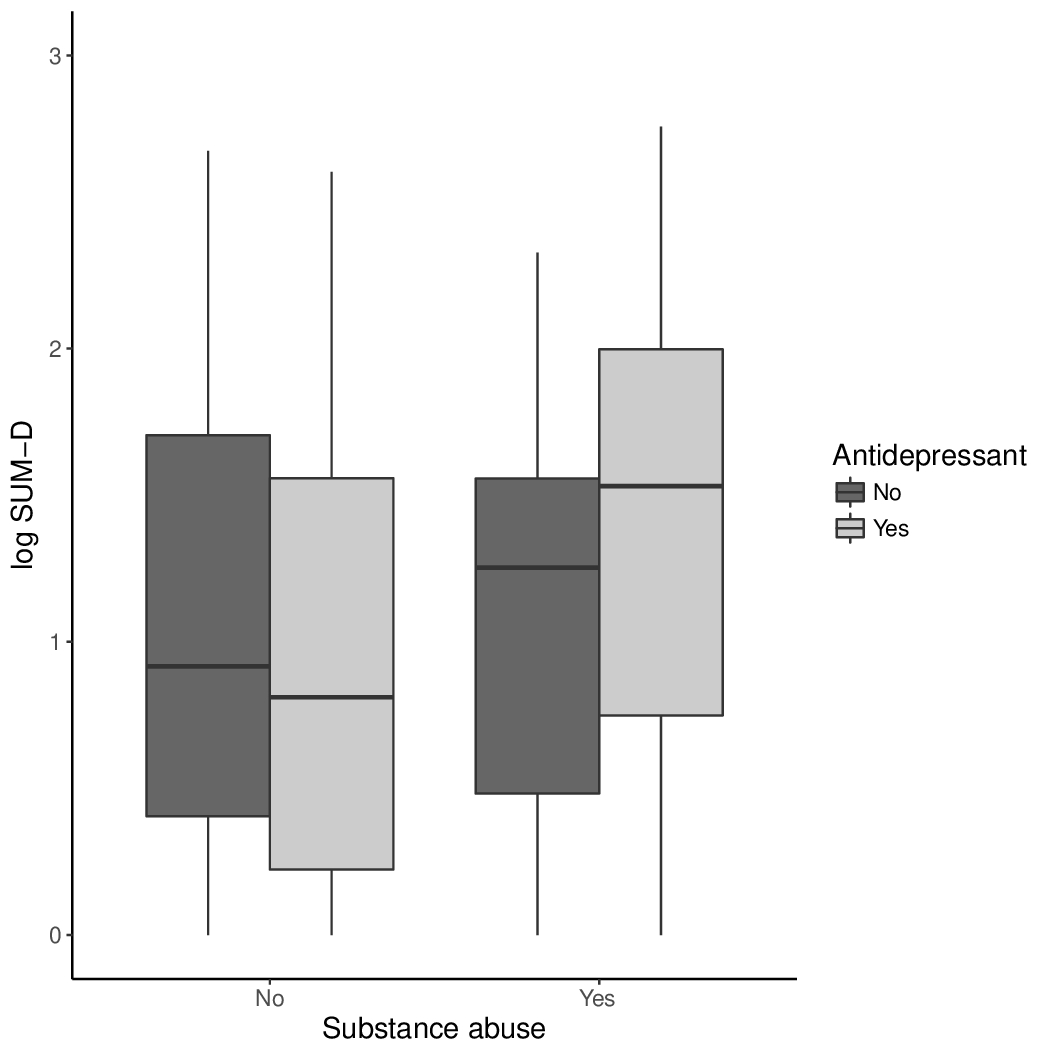}
\caption{Box plots of log SUM-D by substance abuse and treatment.}
\label{sumd_sub_box}
\end{figure}

\begin{figure}[h!]
\centering
\includegraphics[scale = 0.4]{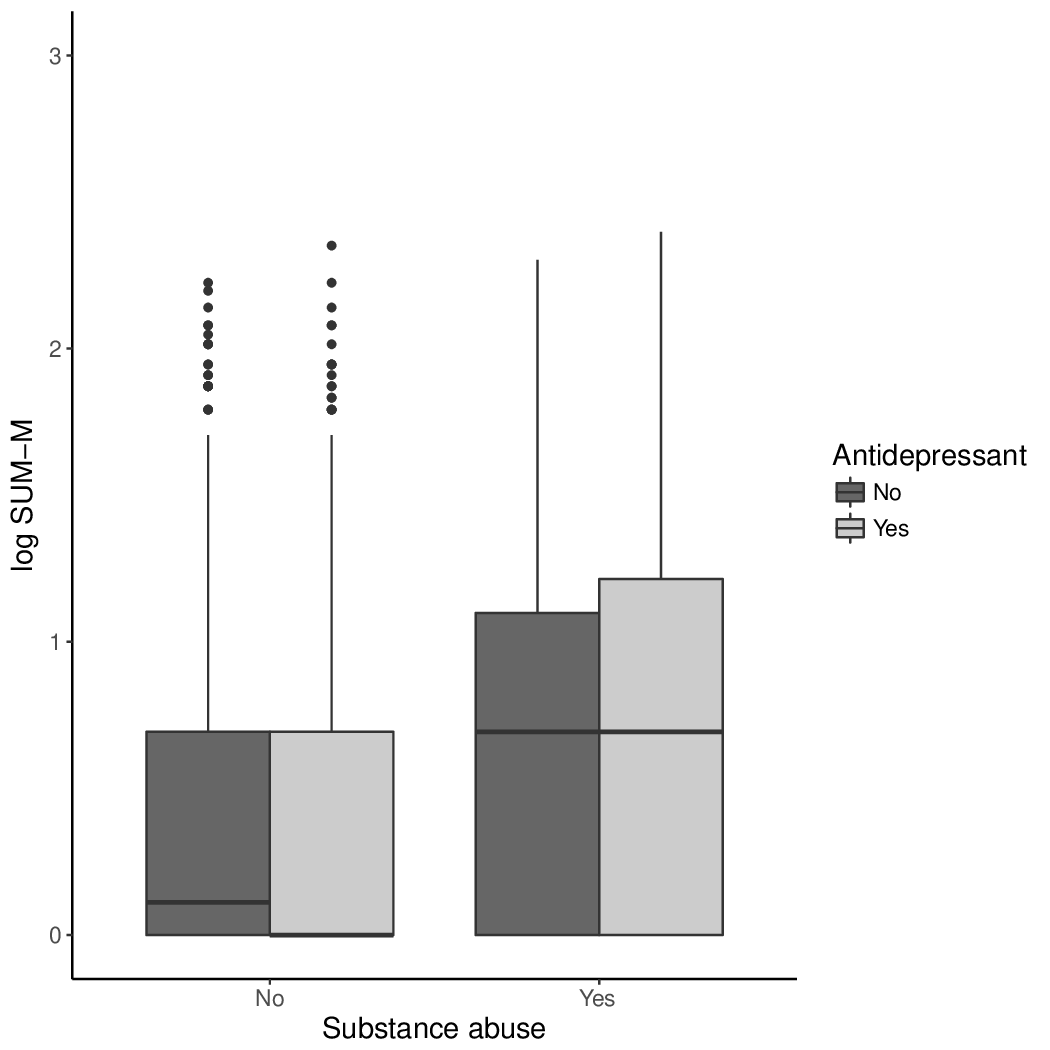}
\caption{Box plots of log SUM-M by substance abuse and treatment.}
\label{summ_sub_box}
\end{figure}

We use the SUM-D score for depression 
symptoms and the SUM-M score for mania symptoms as outcomes. 
We consider a patient treated if they took any one of ten 
antidepressants that appear in the STEP-BD standard care pathway (Deseryl, Serzone, 
Citalopram, Escitalopram Oxalate, Prozac, Fluvoxamine, Paroxetine, Zoloft, Venlafaxine, 
or Bupropion). Covariates used for this analysis were age and substance abuse history (yes/no). 
Figure~\ref{sumd_sub_box} contains box plots of SUM-D scores on the log scale 
by substance abuse and treatment. Figure~\ref{summ_sub_box} contains box plots 
of SUM-M scores on the log scale by substance abuse and treatment. 
For both outcomes, lower scores are more desirable. 
Figure~\ref{sumd_sub_box} indicates that those without a history of substance abuse 
benefit from treatment with antidepressants. However, among those with a history of 
substance abuse, patients treated with antidepressants appear to have worse symptoms of depression. 
Figure~\ref{summ_sub_box} indicates that treatment has no effect on symptoms of mania 
among those without a history of substance abuse. However, among those with a history of 
substance abuse, it appears that treatment may be inducing manic episodes. 
Thus, a sensible treatment policy would be one that tends to prescribe antidepressants only to patients 
without a history of substance abuse. 

We analyzed these data using the proposed method for optimizing composite outcomes. 
Results are summarized in Table~\ref{stepbd_val_tab} below. 
We estimated policies where both utility and probability of optimal treatment 
are fixed (fixed-fixed), where utility is fixed but probability of optimal treatment 
is assumed to vary between patients (fixed-variable), and where both utility and 
probability of optimal treatment are assumed to vary between patients (variable-variable). 
For both the fixed-variable policy and the variable-variable policy, we report 
$\mathbb{E}_n \left\{\mathrm{expit}\left(\bX^\intercal \widehat{\beta}_n\right)\right\}$ 
in place of $\widehat{\rho}_n$ and for the variable-variable policy, we report 
$\mathbb{E}_n \left\{\mathrm{expit}\left(\bX^\intercal \widehat{\theta}_n\right)\right\}$ 
in place of $\widehat{\omega}_n$. Thus, for parameters that are assumed 
to vary across patients, Table~\ref{stepbd_val_tab} contains the mean estimate in the sample. 
Mean outcomes and value functions are averaged over five fold cross validation. 
For both SUM-D and SUM-M, lower scores are preferred. Value is reported as the percent 
improvement over standard of care, calculated using the estimated utility function. 
Large percent improvements in value are preferred.
\begin{table}[h!]
\centering
\begin{tabular}{l|ccccc}
   \hline
Policy & SUM-D & SUM-M & Value (\% improvement) & $\widehat{\omega}_n$ & $\widehat{\rho}_n$ \\ 
   \hline
fixed-fixed & 2.351 & 0.867 & 0.1\% & 0.115 & 0.403 \\ 
  fixed-variable & 2.315 & 0.840 & 3.1\% & 0.115 & 0.405 \\ 
  variable-variable & 2.297 & 0.838 & 7.1\% & 0.173 & 0.405 \\ 
   \hline
standard of care & 2.480 & 0.868 & 0.0\% & $\cdot$ & $\cdot$ \\ 
   \hline
\end{tabular}
\caption{Results of analysis of STEP-BD data for SUM-D and SUM-M.} 
\label{stepbd_val_tab}
\end{table}
All estimated policies produce more desirable SUM-D scores and SUM-M scores 
compared to standard of care and improved value according to the estimated utility. 
Allowing the probability of optimal treatment to vary between patients leads 
to further improvements in value, as does allowing the utility function to 
vary between patients. All policies produce similar estimates for the probability 
of optimal treatment averaged across patients. 

The resulting decision rules can be written as the sign of a linear combination of the covariates. 
As an example, the fixed-fixed policy assigns treatment with antidepressants when 
$\mathrm{sign}\left\{0.207 - 0.003 (\mathrm{age}) - 0.620 (\mathrm{substance \: abuse}) \right\}$ is equal to 1. 
The negative coefficient for substance abuse means that a history of substance abuse 
indicates that a patient should not be prescribed antidepressants. 
Prior research has shown that patients with 
a history of substance abuse are more likely to abuse antidepressants \citep[][]{evans2014abuse}. 
This may contribute to the poor outcomes experienced by STEP-BD patients with a 
history of substance abuse who were treated with antidepressants. 
The coefficients in $\widehat{\theta}_n$ in the variable-variable policy were 
$-0.007$ for age and $7.582$ for substance abuse, with an intercept of $-1.551$. 
A test for preference heterogeneity based on 1000 bootstrap 
samples generated according to Theorem~\ref{bootstrap} yielded a p-value $<0.001$.

As a secondary analysis, we use the SUM-D score and a side effect score as outcomes. 
Eight side effects were recorded 
in the STEP-BD standard care pathway (tremors, dry mouth, sedation, constipation, 
diarrhea, headache, poor memory, sexual dysfunction, and increased appetite). 
Patients rated the severity of each side effect from 0 to 4 with larger values indicating 
more severe side effects. We took the mean score across side effects as the second outcome. 
Patients in the standard care pathway 
were asked to report the percent of days over the past week that they 
experienced mood elevation, irritability, and anxiety; 
these were included as covariates along with age and history of substance abuse. 
Results are summarized in Table~\ref{stepbd_val_tab_2}, reported analogously 
to those in Table~\ref{stepbd_val_tab}. 
\begin{table}[h!]
\centering
\begin{tabular}{l|ccccc}
   \hline
Policy & SUM-D & Side effect score & Value (\% improvement) & $\widehat{\omega}_n$ & $\widehat{\rho}_n$ \\ 
   \hline
fixed-fixed & 2.341 & 0.153 & 5.6\% & 0.798 & 0.460 \\ 
  fixed-variable & 2.393 & 0.150 & 6.9\% & 0.034 & 0.482 \\ 
  variable-variable & 2.400 & 0.154 & 9.3\% & 0.239 & 0.460 \\ 
   \hline
standard of care & 2.481 & 0.172 & 0.0\% & $\cdot$ & $\cdot$ \\ 
   \hline
\end{tabular}
\caption{Results of analysis of STEP-BD data for SUM-D and Side effect score.} 
\label{stepbd_val_tab_2}
\end{table}
Each estimated policy produces improved SUM-D scores and improved side effect scores 
compared to the standard of care. 
Each policy also produces improvement in value according to the estimated 
utility function. Again, allowing the utility function to vary between patients 
results in further improvements in value. 
Each policy produces similar estimates of the probability that patients are 
treated optimally in standard care. The variable-variable policy places more weight 
on SUM-D scores on average compared to the other policies. 
Estimated coefficients in $\widehat{\theta}_n$ are 
$-2.851$ for age, $-2.088$ for substance abuse history, $-1.159$ for 
percent of days with mood elevation, $2.529$ for percent days with irritability, 
and $0.048$ for percent days with anxiety, with an intercept of $0.673$. 
The bootstrap procedure for testing the null hypothesis that patient preferences 
are homogeneous based on 1000 bootstrap samples yielded a p-value $<0.001$.

\section{Discussion}  \label{comp.conclude}

The estimation of individualized treatment rules has been well-studied 
in the statistical literature. Existing methods have typically defined the optimal treatment 
rule as optimizing the mean of a fixed scalar outcome. However, clinical practice often requires 
consideration of multiple outcomes. Thus, there is a disconnect between 
existing statistical methods and current clinical practice. It is reasonable to assume 
that clinicians make treatment decisions for each patient with the goal of maximizing 
that patient's utility. Therefore, it is natural to use observational data to estimate 
patient utilities from observed clinician decisions. This represents a new paradigm for 
thinking about observational data, 
as traditional approaches to analyzing observational data seek transformations 
under which the data can be treated like a randomized study. 

The proposed methodology offers many opportunities for future research. In this paper, 
we have considered only the simplest case--- that of one decision time, two outcomes, 
and two possible treatments. Scenarios with more than two outcomes are discussed in 
the Appendix, and the simulation results there demonstrate that the proposed method 
performs well with three outcomes. Extensions to more than two treatments or multiple 
time points represent potential areas for future research. The proposed method 
requires positing a parametric model for the utility function. Model misspecification 
is discussed in the Appendix, and the simulation results there demonstrate that the 
proposed method performs reasonably well when important covariates are omitted from 
the model for the utility function. Further study of model misspecification represents 
another area for future research. 
Different computational strategies, such as a Bayesian approach, could be developed to 
handle the non-smooth pseudo-likelihood and potentially improve estimation of 
patient-specific utility functions. 
While we have proposed our utility function estimator inside the framework of one-stage 
Q-learning, the pseudo-likelihood utility function estimator could be used alongside other existing 
one-stage optimal treatment policy estimators, such as an outcome weighted learning estimator 
\citep[][]{zhao2012estimating}. There is a great future for the development of methods for 
optimizing composite outcomes in precision medicine and application of these methods in clinical studies. 


\acks{We would like to acknowledge support for this project
from the National Science Foundation (NSF grants DMS-1555141 and DMS-1513579)
and the National Institutes of Health (NIH grants R01 DK108073 and P01 CA142538).  
We also gratefully acknowledge the National Institute of Mental Health for providing access 
to the STEP-BD data set.  }


\newpage

\appendix

\section*{Appendix A: Proofs}
\label{app:theorem}

\begin{proof}[Proof of Theorem~\ref{thm.patient.cons}]
The log of the pseudo-likelihood is given by 
$$
\widehat{\ell}_n(\theta, \beta) = \mathbb{E}_n \left[ \bX^\intercal \beta 
1\left\{A = \widehat{d}_{\theta, n}(\bX) \right\} - \log \left\{1 + 
\exp\left(\bX^\intercal \beta \right) \right\} \right].
$$
Let $\widehat{m}(\cdot, \cdot; \theta, \beta): \mathcal{X} \times \mathcal{A} \rightarrow \mathbb{R}$ 
be defined by $\widehat{m}(\bX, A; \theta, \beta) = \bX^\intercal \beta 
1\left\{A = \widehat{d}_{\theta, n}(\bX) \right\} - \log \left\{1 + 
\exp\left(\bX^\intercal \beta \right) \right\}$ and consider the 
class of functions $\left\{ \widehat{m}(\cdot, \cdot; \theta, \beta) : 
\theta \in \mathbb{R}^p, \beta \in \mathcal{B}  \right\}$. 
The class $\left\{ \log\left\{ 1 + \exp(\bX^\intercal \beta) \right\} : 
\beta \in \mathcal{B} \right\}$
is contained in a VC class by Lemma~9.9 (viii) and (v) 
of \cite{kosorok2008introduction}. 
By Theorem~9.3 of \cite{kosorok2008introduction}, this is 
also a Glivenko--Cantelli (GC) class. 

Let $u(\bX, A; \theta) = \omega(\bX; \theta) 
\left\{\widehat{Q}_{Y, n}(\bX, A) - \widehat{Q}_{Z, n}(\bX, A)\right\} - \widehat{Q}_{Z, n}(\bX, A)$, 
which lies in a VC class indexed by $\theta \in \mathbb{R}^p$ by Lemma~9.6 and Lemma~9.9 (viii), (vi), and (v) 
of \cite{kosorok2008introduction}. We have that 
\begin{multline*}
1\left\{A = \widehat{d}_{\theta, n}(\bX)\right\} = 1(A = 1)1\left\{u(\bX, 1; \theta) - u(\bX, -1, \theta) \ge 0\right\} 
 \\ + 1(A = -1)1\left\{u(\bX, 1; \theta) - u(\bX, -1, \theta) < 0\right\},
\end{multline*}
and it follows that 
$1\left\{A = \widehat{d}_{\theta, n}(\bX)\right\}$ is contained in a GC class 
indexed by $\theta \in \mathbb{R}^p$. 
From Corollary~9.27 (ii) of \cite{kosorok2008introduction} it follows that 
$\bX^\intercal \beta 1\left\{A = \widehat{d}_{\theta, n}(\bX)\right\}$ lies 
in a GC class indexed by $(\theta, \beta) \in \mathbb{R}^p \times \mathcal{B}$ 
as long as $\bX^\intercal \beta$ is uniformly bounded by a function with finite mean, which 
holds as long as $\mathcal{B}$ is compact and $\|\mathbb{E}\bX\| < \infty$.  
It follows that 
$$
\operatorname*{sup}_{(\theta, \beta) \in \mathbb{R}^p \times \mathcal{B}} 
\left| (\mathbb{E}_n - \mathbb{E}) \left[ \bX^\intercal \beta 1\left\{A = \widehat{d}_{\theta, n}(\bX) \right\} 
 - \log\left\{ 1 + \exp(\bX^\intercal \beta)\right\} \right] \right| \xrightarrow[]{P} 0.
$$

Next, define
$$
\widehat{M}(\theta, \beta) = \mathbb{E}\left\{\widehat{m}(\bX, A; \theta, \beta)\right\}
= \mathbb{E}\left( \bX^\intercal \beta\mathbb{E} \left[ 
 1\left\{A = \widehat{d}_{\theta, n}(\bX) \right\} | \bX \right]\right) 
 - \mathbb{E} \log\left\{1 + \mathrm{exp}(\bX^\intercal \beta)\right\}
$$ 
and note that $\widehat{M}(\theta, \beta)$ is continuous in $\beta$. 
The inside expectation of the first piece is 
\begin{multline*}
\mathbb{E} \left[1\left\{A = \widehat{d}_{\theta, n}(\bX)\right\} | \bX\right] 
 = \mathrm{expit}(\bX^\intercal \beta_0) 1\left\{ \widehat{d}_{\theta, n}(\bX) 
 = d^\mathrm{opt}_{\theta_0}(\bX)\right\} \\
 + \left\{1 - \mathrm{expit}(\bX^\intercal \beta_0) \right\}
 1\left\{ \widehat{d}_{\theta, n}(\bX) 
 \ne d^\mathrm{opt}_{\theta_0}(\bX)\right\}, 
\end{multline*}
using the fact that $\mathrm{Pr}\left\{A = d^\mathrm{opt}_{\theta_0}(\bX)\right\} 
= \mathrm{expit}(\bX^\intercal \beta_0)$. 
Define $a(\bX) = Q_Y(\bX, 1) - Q_Y(\bX, -1) - Q_Z(\bX, 1) + Q_Z(\bX, -1)$ 
and $b(\bX) = Q_Z(\bX, 1) - Q_Z(\bX, -1)$. 
Similarly, define $\widehat{a}(\bX) = \widehat{Q}_{Y, n}(\bX, 1) - \widehat{Q}_{Y, n}(\bX, -1) 
 - \widehat{Q}_{Z, n}(\bX, 1) + \widehat{Q}_{Z, n}(\bX, -1)$ 
and $\widehat{b}(\bX) = \widehat{Q}_{Z, n}(\bX, 1) - \widehat{Q}_{Z, n}(\bX, -1)$.
Then, 
\begin{eqnarray*}
1\left\{ \widehat{d}_{\theta, n}(\bX) = d^\mathrm{opt}_{\theta_0}(\bX) \right\} 
 & = & 1\left[ \left\{\omega(\bX; \theta) \widehat{a}(\bX) + \widehat{b}(\bX)\right\}
 \left\{ \omega(\bX; \theta) a(\bX) + b(\bX)\right\} \ge 0\right] \\
 & = & 1\Big[ \omega(\bX; \theta) 
 \left\{ \omega(\bX; \theta) a(\bX)\widehat{a}(\bX) + \widehat{a}(\bX) b(\bX) \right\} \\
 &  & + \omega(\bX; \theta) a(\bX) \widehat{b}(\bX) + \widehat{b}(\bX) b(\bX) \ge 0\Big],
\end{eqnarray*}
and thus $\mathbb{E} \left[1\left\{A = \widehat{d}_{\theta, n}(\bX) \right\} | \bX \right]$ 
is continuous in $\theta$. 

Let $m(\bX, A; \theta, \beta) = \bX^\intercal \beta 
1\left\{A = d^\mathrm{opt}_\theta(\bX) \right\} - \log \left\{1 + 
\exp\left(\bX^\intercal \beta \right) \right\}$. 
Because the model is identifiable and $\mathcal{L}_n(\theta, \beta)$ is a parametric log-likelihood, 
$\mathbb{E} m(\bX, A; \theta, \beta)$ has unique maximizers at $\theta_0$ and $\beta_0$. 
Let $\widetilde{\theta}$ and $\widetilde{\beta}$ be the maximizers of 
$\mathbb{E}\widehat{m}(\bX, A; \theta, \beta)$. Because 
$\mathbb{E}\left\{ \widehat{d}_{\theta, n}(\bX) | \bX = \bx\right\}$ $ = d^\mathrm{opt}_\theta(\bx)$ 
for any $\bx \in \mathcal{X}$ and $\theta \in \mathbb{R}^p$, 
$\mathbb{E}\left[1\left\{A = \widehat{d}_{\theta, n}(\bX)\right\} | \bX\right] 
 = \mathbb{E}\left[1\left\{A = d^\mathrm{opt}_\theta(\bX)\right\} | \bX\right]$, 
which implies that $\mathbb{E} m(\bX, A; \theta, \beta) = \mathbb{E} \widehat{m}(\bX, A; \theta, \beta)$. 
Thus, $\widetilde{\theta} = \theta_0$ and $\widetilde{\beta} = \beta_0$. 
The claim now follows from Lemma~14.3 and Theorem~2.12 of \cite{kosorok2008introduction}. 
\end{proof}

\begin{proof}[Proof of Theorem~\ref{thm.patient.value}]
Define $Q_{\theta_0}(\bx ,a)$ and $Q_{\widehat{\theta}_n}(\bx, a)$ 
as defined in Section~\ref{comp.single}. Let $u(Y, Z; A, \bX, \theta) = 
\omega(\bX; \theta) Q_Y(\bX, A) + 
\left\{ 1 - \omega(\bX; \theta)\right\} Q_Z(\bX, A)$. 
Under the given assumptions, for some constant $0 < c < \infty$,   
\begin{multline} \label{qian3.1}
\left| V\left(\widehat{d}_{\widehat{\theta}_n, n}\right) 
 - V\left(d^\mathrm{opt}_{\theta_0}\right) \right| \\
 \le c \left| \mathbb{E}\left\{u(Y, Z; A, \bX, \theta_0) - \widehat{Q}_{\widehat{\theta}_n, n}(\bX, A)\right\}^2 
 - \mathbb{E}\left\{u(Y, Z; A, \bX, \theta_0) - Q_{\theta_0}(\bX, A)\right\}^2 \right|^{1/2}
\end{multline}
by equation~(3.1) of \cite{qian2011performance} 
\citep[see also][]{murphy2005generalization}. 
The right hand side of (\ref{qian3.1}) converges in probability to 0 
by the consistency of $\widehat{\theta}_n$, consistency of $\widehat{Q}_{Y, n}$ 
and $\widehat{Q}_{Z, n}$, and the continuous mapping theorem. 
The result follows. 
\end{proof}

\begin{proof}[Proof of Theorem~\ref{asymptotic.distribution}]
By definition of $\left(\widehat{\theta}_n, \widehat{\beta}_n\right)$, 
\begin{eqnarray*}
0 & \le & \widehat{\ell}_n\left(\widehat{\theta}_n, \widehat{\beta}_n\right) 
  - \widehat{\ell}_n\left(\theta_0, \beta_0\right) \\
 & = & \sum_{i = 1}^n \Big[ \bX_i^\intercal \widehat{\beta}_n 1\left\{A_i = \widehat{d}_{\widehat{\theta}_n, n}(\bX_i)\right\} 
  - \bX_i^\intercal \beta_0 1\left\{A_i = \widehat{d}_{\theta_0, n}(\bX_i)\right\} \\
 &  & - \left(\widehat{\beta}_n - \beta_0\right)^\intercal \bX_i P_{\beta_0}(\bX_i) \Big]
  - \frac{1}{2} \sqrt{n} \left(\widehat{\beta}_n - \beta_0\right)^\intercal I_n(\beta_*) 
   \sqrt{n} \left(\widehat{\beta}_n - \beta_0\right) \\
 & = & \sqrt{n} \left(\widehat{\beta}_n - \beta_0\right)^\intercal n^{-1/2} 
  \sum_{i = 1}^n \bX_i \left[ 1\left\{A_i = \widehat{d}_{\widehat{\theta}_n, n}(\bX_i)\right\} - P_{\beta_0}(\bX_i) \right] \\
 &  & - \frac{1}{2} \sqrt{n} \left(\widehat{\beta}_n - \beta_0\right)^\intercal I_n(\beta_*) 
 	\sqrt{n} \left(\widehat{\beta}_n - \beta_0\right) \\
 &  & + \sum_{i = 1}^n \bX_i^\intercal \beta_0 \left[ 1\left\{A_i = \widehat{d}_{\widehat{\theta}_n, n}(\bX_i) \right\} 
	 - 1\left\{A_i = \widehat{d}_{\theta_0, n}(\bX_i)\right\} \right],
\end{eqnarray*}
where $\beta_*$ is a point between $\widehat{\beta}_n$ and $\beta_0$. 
Using the definition of a maximizer and letting 
$\widehat{u}_n(\theta) = n^{-1/2}\sum_{i = 1}^n \bX_i \left[1\left\{A_i = \widehat{d}_{\theta, n}(\bX_i)\right\} 
 - P_{\beta_0}(\bX_i)\right]$, we have that 
$\sqrt{n}\left(\widehat{\beta}_n 
 - \beta_0\right) = I_n(\beta_*)^{-1} \widehat{u}_n\left(\widehat{\theta}_n\right)$,
since $I_n(\beta_*) \xrightarrow[]{P} I_0$ and $I_0$ is positive definite. Next, note that 
%
%
\begin{eqnarray*}
\widehat{u}_n\left(\widehat{\theta}_n\right) & = & n^{-1/2} \sum_{i = 1}^n \bX_i 
 \left[1\left\{A_i = \widehat{d}_{\widehat{\theta}_n, n}(\bX_i)\right\} 
 - 1\left\{A_i = d^\mathrm{opt}_{\theta_0}(\bX_i)\right\}\right] + Z_{A, n} \\ 
 & = & \mathbb{G}_n \left( \bX \left[1\left\{A = \widehat{d}_{\widehat{\theta}_n, n}(\bX)\right\} 
 - 1\left\{A = d^\mathrm{opt}_{\theta_0}(\bX)\right\}\right] \right) + Z_{A, n} \\ 
 &  & + \sqrt{n} \mathbb{E}\left( \bX \left[ 1\left\{A = \widehat{d}_{\widehat{\theta}_n, n}(\bX)\right\} 
 - 1\left\{A = d^\mathrm{opt}_{\theta_0}(\bX)\right\} \right] \right) \\
 & = & Z_{A, n} + \sqrt{n} 
\mathbb{E}\left( \bX \left[ 1\left\{A = \widehat{d}_{\widehat{\theta}_n, n}(\bX)\right\} 
 - 1\left\{A = d^\mathrm{opt}_{\theta_0}(\bX)\right\} \right] \right) \left\{1 + o_P(1)\right\},  
\end{eqnarray*}
where $\mathbb{G}_n f = n^{1/2} (\mathbb{E}_n - \mathbb{E})f(\bX)$ and $Z_{A,n}=n^{-1/2}\sum_{i=1}^{n}\left[
1\left\{A_i = d^\mathrm{opt}_{\theta_0}(\bX_i)\right\}-P_{\beta_0}(\bX_i)\right]$. We also have 
$1\left\{A = \widehat{d}_{\widehat{\theta}_n, n}(\bX)\right\} - 1\left\{A = d^\mathrm{opt}_{\theta_0}(\bX)\right\}
 = -\left[2 \cdot 1\left\{A = d^\mathrm{opt}_{\theta_0}(\bX)\right\} - 1\right]$ \newline
 $\cdot 1\left\{ \widehat{d}_{\widehat{\theta}_n, n}(\bX) \ne d^\mathrm{opt}_{\theta_0}(\bX) \right\}$,
which implies that 
\begin{eqnarray*}
 &  & \sqrt{n} \mathbb{E}\left(\bX \left[ 1\left\{A = \widehat{d}_{\widehat{\theta}_n, n}(\bX)\right\} 
  - 1\left\{A = d^\mathrm{opt}_{\theta_0}(\bX)\right\} \right] \right) \\
 &  & \hspace{0.2in} = -\sqrt{n} \mathbb{E} \left[\bX \left\{2 P_{\beta_0}(\bX) - 1\right\} 
  1\left\{ \widehat{d}_{\widehat{\theta}_n, n}(\bX) \ne d^\mathrm{opt}_{\theta_0}(\bX) \right\} \right] \\
 & & \hspace{0.2in} = -\sqrt{n} \mathbb{E}\Big\{ \bX\left\{2 P_{\beta_0}(\bX) - 1\right\} 
 \Big( 1\left[0 \le D_{\theta_0}(\bX) < -\left\{\widehat{D}_{\widehat{\theta}_n, n}(\bX) - D_{\theta_0}(\bX)\right\}\right] \\
 &  & \hspace{0.35in} + 1\left[ - \left\{\widehat{D}_{\widehat{\theta}_n, n}(\bX) - D_{\theta_0}(\bX)\right\} 
 \le D_{\theta_0}(\bX) < 0\right] \Big) \Big\} \\
 &  & \hspace{0.2in} = -\mathbb{E}\left[ \bX\left\{2P_{\beta_0}(\bX) - 1\right\} \cdot 
 \left| \sqrt{n} \left\{ \widehat{D}_{\widehat{\theta}_n, n}(\bX) 
 - D_{\theta_0}(\bX) \right\} \right| \Big| D_{\theta_0}(\bX) = 0 \right] f_0 + o_P(1)
\end{eqnarray*}
by Assumption~\ref{assume.B} and the fact that 
$\left\| \widehat{D}_{\widehat{\theta}_n, n}(\bx) - D_{\theta_0}(\bx) \right\|_\mathcal{X} = o_P(1)$. 
Note that 
\begin{eqnarray*}
\sqrt{n} \left\{ \widehat{D}_{\widehat{\theta}_n, n}(\bX) - D_{\theta_0}(\bX)\right\} 
 & = & \sqrt{n} \bigg[ \omega\left(\bX; \widehat{\theta}_n\right) \left\{\widehat{R}_{Y, n}(\bX) - R_Y(\bX) \right\} \\
 &  & + \left\{1 - \omega\left(\bX; \widehat{\theta}_n\right)\right\} \left\{\widehat{R}_{Z, n}(\bX) - R_Z(\bX) \right\} \bigg] \\
 &  & + \sqrt{n} \left\{\omega\left(\bX; \widehat{\theta}_n\right) - \omega\left(\bX; \theta_0\right)\right\}
 \left\{R_Y(\bX) - R_Z(\bX)\right\} \\
 & = & \omega\left(\bX; \theta_0\right) \phi_Y(\bX)^\intercal Z_{Y, n} 
 + \left\{1 - \omega\left(\bX; \theta_0\right)\right\} \phi_Z(\bX)^\intercal Z_{Z, n} \\
 &  & + \dot{\omega}_{\theta_0}(\bX) \left\{R_Y(\bX) - R_Z(\bX)\right\} \sqrt{n} \left(\widehat{\theta}_n - \theta_0\right)
 \left\{1 + o_P(1)\right\} \\
 & = & O_P\left(1 + \sqrt{n}\left\| \widehat{\theta}_n - \theta_0\right\|\right), 
\end{eqnarray*}
thus, $\left\| \widehat{u}_n\left(\widehat{\theta}_n\right)\right\| 
= O_P\left(1 + \sqrt{n}\left\| \widehat{\theta}_n - \theta_0\right\|\right)$. 
Letting $v_n\left(\widehat{\theta}_n, \beta_*\right) = 
n^{-1/2} \widehat{u}_n\left(\widehat{\theta}_n\right)^\intercal I_n(\beta_*)$ 
$\cdot\widehat{u}_n\left(\widehat{\theta}_n\right)$, 
\begin{eqnarray*}
0 & \le & n^{-1/2} \left\{ \widehat{\ell}_n\left(\widehat{\theta}_n, \widehat{\beta}_n\right)
  - \widehat{\ell}_n\left(\theta_0, \beta_0\right) \right\} \\ 
 & = & n^{-1/2} \sum_{i = 1}^n \bX_i^\intercal \beta_0 \left[ 1\left\{A_i = \widehat{d}_{\widehat{\theta}_n, n}(\bX)\right\}
  - 1\left\{A_i = \widehat{d}_{\theta_0, n}(\bX)\right\} \right] 
	+ v_n\left(\widehat{\theta}_n, \beta_*\right)/2 \\
 & = & n^{1/2} \mathbb{E}\left( \bX^\intercal \beta_0 \left[1\left\{A = \widehat{d}_{\widehat{\theta}_n, n}(\bX)\right\} 
  - 1\left\{A = \widehat{d}_{\theta_0, n}(\bX)\right\} \right] \right) 
  + o_P\left(1 + \sqrt{n}\left\|\widehat{\theta}_n - \theta_0\right\|\right) \\
 & = & n^{1/2} \mathbb{E}\left( \bX^\intercal \beta_0 \left[1\left\{A = \widehat{d}_{\widehat{\theta}_n, n}(\bX)\right\} 
  - 1\left\{A = d^\mathrm{opt}_{\theta_0}(\bX)\right\} \right] \right) \\
 &  & - n^{1/2} \mathbb{E}\left( \bX^\intercal \beta_0 \left[1\left\{A = \widehat{d}_{\theta_0, n}(\bX)\right\} 
  - 1\left\{A = d^\mathrm{opt}_{\theta_0}(\bX)\right\} \right] \right) 
  + o_P\left(1 + \sqrt{n}\left\|\widehat{\theta}_n - \theta_0\right\|\right) \\
 & = & \mathbb{E} \left[ \bX^\intercal \beta_0 \left\{ 2 P_{\beta_0}(\bX) - 1\right\} \cdot 
  \left| \sqrt{n} \left\{\widehat{D}_{\widehat{\theta}_n, n}(\bX) - D_{\theta_0}(\bX)\right\} \right| 
	\Big| D_{\theta_0}(\bX) = 0\right] f_0 r_n \\ 
 &  & + O_P(1) + o_P\left(1 + \sqrt{n} \left\|\widehat{\theta}_n - \theta_0\right\|\right) \\
 & \le & -\mathbb{E} \left[ \bX^\intercal \beta_0 \left\{ 2 P_{\beta_0}(\bX) - 1\right\} \cdot 
  \left| \left\{R_Y(\bX) - R_Z(\bX)\right\} \dot{\omega}_{\theta_0}(\bX)^\intercal \left(\widehat{\theta}_n - \theta_0\right) \right| 
	\Big| D_{\theta_0}(\bX) = 0 \right] f_0 r_n \\ 
 &  & + O_P(1) + o_P\left(1 + \sqrt{n} \left\|\widehat{\theta}_n - \theta_0\right\|\right) \\
 & \le & -\delta_2 \delta_1^2 \left(\frac{\mathrm{exp}(\delta_1) - 1}{\mathrm{exp}(\delta_1) + 1}\right) \sqrt{n}
 \left\| \widehat{\theta}_n - \theta_0 \right\| \left\{1 + o_P(1)\right\} + O_P(1) 
	+ o_P\left(1 + \sqrt{n} \left\|\widehat{\theta}_n - \theta_0\right\|\right), 
\end{eqnarray*}
where $r_n = 1 + o_P(1)$. This 
implies that $\sqrt{n}\left\| \widehat{\theta}_n - \theta_0\right\| = O_P(1)$. 
Let 
\begin{multline*}
M_n\left(\widehat{\theta}_n\right) = n^{-1/2} \sum_{i = 1}^n \bX_i^\intercal \beta_0 
 \left[ 1\left\{A_i = \widehat{d}_{\widehat{\theta}_n, n}(\bX)\right\}
 - 1\left\{A_i = \widehat{d}_{\theta_0, n}(\bX)\right\} \right] 
 + v_n\left(\widehat{\theta}_n, \beta_*\right)/2,
\end{multline*}
let $M(u) = \beta_0^\intercal k_0(Z_Y, Z_Z, u)$, and let $U = \mathrm{arg \, min}_{u \in \mathbb{R}^d} M(u)$. 
We will show that $M_n(\theta_0 + u / \sqrt{n}) \rightsquigarrow M(u)$ in $\ell^\infty(K)$ for any compact 
subset $K$ of $\mathbb{R}^d$. Then, it will follow from the argmax Theorem \citep[See chapter 14 of ][]{kosorok2008introduction} 
that $\widetilde{U}_n \rightsquigarrow U$, 
where $\widetilde{U}_n = \mathrm{arg \, min}_{u \in \mathbb{R}^d} M_n(\theta_0 + u / \sqrt{n})$. 
Let $h_n(u) = \theta_0 + u / \sqrt{n}$. 
Similar arguments along with Assumptions~\ref{assume.C} and~\ref{assume.B}, yield that, for any 
compact $K \subset \mathbb{R}^d$, 
\begin{eqnarray*}
\lefteqn{\operatorname*{arg \, min}_{u \in K} M_n\left\{h_n(u)\right\}}&&\\
&=&\operatorname*{arg \, min}_{u \in K}
  n^{1/2} \mathbb{E} \left( \bX^\intercal \beta_0 \left[ 1\left\{A = \widehat{d}_{h_n(u), n}(\bX) \right\} 
  - 1\left\{A = d^\mathrm{opt}_{\theta_0} (\bX) \right\}\right] \right) \\ 
 & = & \operatorname*{arg \, min}_{u \in K} n^{1/2} \mathbb{E} \Big\{ \bX^\intercal \beta_0 
  \left\{ 2 P_{\beta_0}(\bX) - 1 \right\} \Big( 1\Big[ - \left\{ \widehat{D}_{h_n(u), n}(\bX) 
  - D_{\theta_0}(\bX)\right\} \\
	&  & \le D_{\theta_0}(\bX) < 0\Big] + 1\left[ 0 \le D_{\theta_0}(\bX) < 
  -\left\{\widehat{D}_{h_n(u), n}(\bX) - D_{\theta_0}(\bX)\right\} \right] \Big) \Big\} \\
 & = & \operatorname*{arg \, min}_{u \in K} \mathbb{E} \Big[ \bX^\intercal \beta_0 
  \left\{ 2 P_{\beta_0}(\bX) - 1\right\} \left| \sqrt{n} \left\{ \widehat{D}_{h_n(u), n}(\bX) 
	- D_{\theta_0}(\bX)\right\} \right| \\
 &  & \Big| D_{\theta_0}(\bX) = 0 \Big] f_0 + o_P(1),
\end{eqnarray*}
However, 
\begin{multline*}
\sqrt{n} \left\{ \widehat{D}_{h_n(u), n}(\bX) - D_{\theta_0}(\bX)\right\} \rightsquigarrow 
 \omega\left(\bX; \theta_0\right) \phi_Y(\bX)^\intercal Z_Y
 + \left\{ 1 - \omega\left(\bX; \theta_0\right)\right\} \phi_Z(\bX)^\intercal Z_Z \\
 + R_0(\bX) \dot{\omega}_{\theta_0}(\bX)^\intercal u
\end{multline*}
uniformly over $\mathcal{X}$ when $\bX$ has its conditional distribution given $D_{\theta_0}(\bX) = 0$. 
This implies that $M_n(\theta + u / \sqrt{n}) \rightsquigarrow M(u)$ in $\ell^\infty(K)$ as desired and 
thus $\widetilde{U}_n \rightsquigarrow U$. It is straightforward to verify the remaining conclusions of the theorem 
using previous arguments. 
\end{proof}

\begin{proof}[Proof of Theorem~\ref{bootstrap}]
Using the assumptions, the fact that both $\sqrt{n}\left(\widehat{\theta}_n - \theta_0\right) = O_P(1)$ and 
$\sqrt{n}\left(\widehat{\beta}_n - \beta_0\right) = O_P(1)$, and standard arguments, we obtain that, 
for any compact $K_1 \subset \mathbb{R}^q$, 
$\mathrm{sup}_{(Z_Y^\intercal, Z_Z^\intercal)^\intercal \in K_1} 
\left\| \widetilde{T}_n\left\{ \bx, \widetilde{Z}_Y(Z_Y), \widetilde{Z}_Z(Z_Z)\right\} - T_0(\bx, Z_Y, Z_Z)\right\|_\mathcal{X}
= o_P(1)$, where $\widetilde{Z}_Y(Z_Y) = \widehat{\Sigma}_n^{1/2} \Sigma_0^{-1/2} (Z_Y^\intercal, 0^\intercal, 0^\intercal)^\intercal$, 
$\widetilde{Z}_Z(Z_Z) = \widehat{\Sigma}_n^{1/2} \Sigma_0^{-1/2} (0^\intercal, Z_Z^\intercal, 0^\intercal)^\intercal$, 
and also \newline 
$T_0(\bx, Z_Y, Z_Z) = \omega(\bx; \theta_0) \phi_Y(\bx)^\intercal Z_Y + \left\{1 - \omega(\bx; \theta_0)\right\} 
\phi_Z(\bx)^\intercal Z_Z$. Furthermore, 
\begin{eqnarray*}
 &  & \left\| \left\{ \widehat{R}_{Y, n}(\bx) - \widehat{R}_{Z, n}(\bx)\right\} \dot{\omega}_{\widehat{\theta}_n}(\bx)^\intercal u 
 - \left\{ R_Y(\bx) - R_Z(\bx)\right\} \dot{\omega}_{\theta_0}(\bx)^\intercal u \right\|_\mathcal{X} \\
 &  & \hspace{0.2in} \le \bigg\| \left\| \left\{ \widehat{R}_{Y, n}(\bx) - \widehat{R}_{Z, n}(\bx)\right\} 
	\dot{\omega}_{\widehat{\theta}_n}(\bx) 
  - \left\{ R_Y(\bx) - R_Z(\bx)\right\} \dot{\omega}_{\theta_0}(\bx) \right\| \bigg\|_\mathcal{X} \cdot \|u\| \\
 &  & \hspace{0.2in} = O_P\left(n^{-1/2}\right) \|u\|,
\end{eqnarray*}
$\left\| \widehat{D}_{\widehat{\theta}_n, n}(\bx) - D_{\theta_0}(\bx) \right\|_\mathcal{X} = O_P\left(n^{-1/2}\right)$, 
and $\left\| P_{\widehat{\beta}_n}(\bx) - P_{\beta_0}(\bx) \right\|_\mathcal{X} = O_P\left(n^{-1/2}\right)$. 
Thus, 
\begin{multline} \label{boot.eq.1}
\operatorname*{sup}_{(Z_Y^\intercal, Z_Z^\intercal)^\intercal \in K_1} \mathbb{E}_n 
\left[ \|\bX\| \cdot \left| \left\{ 2 P_{\widehat{\beta}_n}(\bX) - 1\right\} 
\widetilde{T}_n\left\{ \bX, \widetilde{Z}_Y(Z_Y), \widetilde{Z}_Z(Z_Z) \right\} \right| \frac{1}{h_n} 
\phi_0\left\{ \frac{\widehat{D}_{\widehat{\theta}_n,n }(\bX)}{h_n} \right\} \right] \\
\le O_P(1) \mathbb{E}_n \left[ \frac{1}{h_n} 
\phi_0\left\{ \frac{\widehat{D}_{\widehat{\theta}_n,n }(\bX)}{h_n} \right\} \right]. 
\end{multline}
However,  
\begin{eqnarray*}
&  & \mathbb{E}_n \left( \frac{1}{h_n} \left[ \phi_0\left\{ \frac{\widehat{D}_{\widehat{\theta}_n,n }(\bX)}{h_n} \right\}
 - \phi_0\left\{ \frac{D_{\theta_0}(\bX)}{h_n} \right\} \right] \right) \\
 &  & \hspace{0.1in} = \mathbb{E}_n \Bigg[ \frac{1}{h_n^3} \int_0^1 \left\{ (1 - s) D_{\theta_0}(\bX) 
 + s \widehat{D}_{\widehat{\theta}_n, n}(\bX) \right\} \phi_0\left\{ \frac{(1 - s) D_{\theta_0}(\bX) 
 + s \widehat{D}_{\widehat{\theta}_n, n}(\bX)}{h_n} \right\} \mathrm{d}s \\
 &  & \hspace{0.2in} \cdot \left\{ \widehat{D}_{\widehat{\theta}_n, n}(\bX) - D_{\theta_0}(\bX)\right\} \Bigg] \\
 &  & \hspace{0.1in} = O_P\left( \frac{1}{h_n^3 n^{1/2}}\right)\\
 && \hspace{0.2in}\cdot\mathbb{E}_n \left[ \int_0^1 \left\{ (1 - s) D_{\theta_0}(\bX) 
 + s \widehat{D}_{\widehat{\theta}_n, n}(\bX) \right\} \phi_0\left\{ \frac{(1 - s) D_{\theta_0}(\bX) 
 + s \widehat{D}_{\widehat{\theta}_n, n}(\bX)}{h_n} \right\} \mathrm{d}s \right] \\
 &  & \hspace{0.1in} = O_P\left( \frac{1}{h_n^3 n^{1/2}}\right) O_P(h_n) 
 = O_P\left( \frac{1}{h_n^3 n^{1/2}}\right) = o_P(1),
\end{eqnarray*}
since $|u\phi_0(u)| \le (2\pi)^{-1/2} e^{-1} < \infty$. Now, since 
$\mathbb{E} \left[ h_n^{-1} \phi_0 \left\{ D_{\theta_0}(\bX) / h_n\right\} \right] \xrightarrow[]{P} f_0$, 
we have that~(\ref{boot.eq.1}) is equal to $O_P(1)$. Thus, if $\left\|u_n\right\| \rightarrow \infty$, 
\begin{multline} \label{boot.eq.2}
\widehat{\beta}_n^\intercal \widetilde{k}_n\left( \widetilde{Z}_Y, \widetilde{Z}_Z, u_n\right) \ge 
 \mathbb{E}_n \Bigg[ \widehat{\beta}_n^\intercal \bX \left\{ 2 P_{\widehat{\beta}_n}(\bX) - 1\right\} 
 \cdot \left| \left\{ \widehat{R}_{Y, n}(\bX) - \widehat{R}_{Z, n}(\bX)\right\} 
 \dot{\omega}_{\widehat{\theta}_n}(\bX)^\intercal u_n \right| 
 \\ \cdot \frac{1}{h_n} \phi_0\left\{ \frac{\widehat{D}_{\widehat{\theta}_n, n}(\bX)}{h_n} \right\} \Bigg] - O_P(1),
\end{multline}
where the $O_P(1)$ is uniform over $K_1$. Thus, up to the $O_P(1)$ added on the right-hand side, 
\begin{eqnarray*}
(\ref{boot.eq.2}) & \ge & \|u_n\| \\
&&\cdot\operatorname*{inf}_{t \in S^d} \mathbb{E}_n \left[ \widehat{\beta}_n^\intercal \bX 
\left\{ 2 P_{\widehat{\beta}_n}(\bX) - 1\right\} \left| \left\{ \widehat{R}_{Y, n}(\bX) - \widehat{R}_{Z, n}(\bX)\right\} 
 \dot{\omega}_{\widehat{\theta}_n}(\bX)^\intercal t \right| \frac{1}{h_n} 
 \phi_0\left\{ \frac{\widehat{D}_{\widehat{\theta}_n, n}(\bX)}{h_n} \right\} \right] \\ 
 & \ge &  \|u_n\| \bigg( o_P(1) + \operatorname*{inf}_{t \in S^d} \mathbb{E} \bigg[ \beta_0^\intercal \bX 
\left\{ 2 P_{\beta_0}(\bX) - 1\right\} \left| \left\{ R_Y(\bX) - R_Z(\bX)\right\} 
 \dot{\omega}_{\theta_0}(\bX)^\intercal t \right| \\
 &  & \frac{1}{h_n} \phi_0\left\{ \frac{D_{\theta_0}(\bX)}{h_n} \right\} \bigg] \bigg) \\ 
 & = & \|u_n\| \\
&&\cdot \left( o_P(1) + \operatorname*{inf}_{t \in S^d} \mathbb{E} \left[ \beta_0^\intercal \bX 
\left\{ 2 P_{\beta_0}(\bX) - 1\right\} \left| \left\{ R_Y(\bX) - R_Z(\bX)\right\} 
 \dot{\omega}_{\theta_0}(\bX)^\intercal t \right| \Big| D_{\theta_0}(\bX) = 0 \right] f_0 \right) \\ 
 & \ge & \|u_n\| \left[ o_P(1) + \delta_2 \delta_1^2 
 \left\{ \frac{\mathrm{exp}(\delta_1) - 1}{\mathrm{exp}(\delta_1) + 1} \right\} \right], 
\end{eqnarray*}
with the expectation over $\bX$. Let 
$
\widehat{U}_n\left(Z_Y, Z_Z\right) 
 = \operatorname*{arg \, min}_{u \in \mathbb{R}^d} \widehat{\beta}_n^\intercal 
 \widetilde{k}_n\left\{ \widetilde{Z}_Y\left(Z_Y\right), \widetilde{Z}_Z\left(Z_Z\right), u\right\},
$ 
where, if the arg min set has more than one element, one can be chosen randomly or algorithmically. 
Since the $O_P(1)$ above is uniform over $K_1$, we conclude that
\begin{equation} \label{boot.eq.4}
\operatorname*{sup}_{(Z_Y^\intercal, Z_Z^\intercal)^\intercal \in K_1} \left\| \widehat{U}_n\left(Z_Y, Z_Z\right)\right\| = O_P(1).
\end{equation}
Now, let $K_2$ be any compact subset of $\mathbb{R}^d$. Previous and standard arguments give us that 
$
\operatorname*{sup}_{(Z_Y^\intercal, Z_Z^\intercal)^\intercal \in K_1} \operatorname*{sup}_{u \in K_2} 
 \left\| \widetilde{k}_n \left\{ \widetilde{Z}_Y\left(Z_Y\right), \widetilde{Z}_Z\left(Z_Z\right), u\right\}
 - k_0\left(Z_Y, Z_Z, u\right) \right\| = o_P(1).
$
Thus, we also have that 
\begin{equation} \label{boot.eq.6}
\operatorname*{sup}_{(Z_Y^\intercal, Z_Z^\intercal)^\intercal \in K_1} \operatorname*{sup}_{u \in K_2} 
 \left\| \widehat{\beta}_n^\intercal \widetilde{k}_n \left\{ \widetilde{Z}_Y\left(Z_Y\right), 
 \widetilde{Z}_Z\left(Z_Z\right), u\right\}
 - \beta_0^\intercal k_0\left(Z_Y, Z_Z, u\right) \right\| = o_P(1).
\end{equation}
Define $U_0(Z_Y, Z_Z) = \mathrm{arg \, max}_{u \in \mathbb{R}^d} \beta_0^\intercal k_0(Z_Y, Z_Z, u)$. Previous arguments yield that 
\begin{equation} \label{boot.eq.5}
\operatorname*{sup}_{(Z_Y^\intercal, Z_Z^\intercal)^\intercal \in K_1} \left\| U_0(Z_Y, Z_Z) \right\| = O(1).
\end{equation} 
By Assumption~\ref{assume.C}, the arg min for each $\left(Z_Y^\intercal, Z_Z^\intercal\right)^\intercal \in K_1$ is unique. 
Fix $\epsilon > 0$. By~(\ref{boot.eq.4}), there exists an $m_2 < \infty$ such that 
$\mathrm{Pr}\left( \mathrm{sup}_{(Z_Y^\intercal, Z_Z^\intercal)^\intercal \in K_1} 
 \left\| \widehat{U}_n\left(Z_Y, Z_Z\right)\right\| < m_2 \right) \ge 1 - \epsilon$ 
for all n large enough. By~(\ref{boot.eq.5}), we can enlarge $m_2$ such that 
$$
\mathrm{sup}_{(Z_Y^\intercal, Z_Z^\intercal)^\intercal \in K_1} \left\| U_0(Z_Y, Z_Z) \right\| < m_2 < \infty.
$$ 
We can also find an $m_1 < \infty$ such that $K_1 \subset K_{m_1}^q$ as defined in Corollary~\ref{cor}. 
It is straightforward to show that~(\ref{cor.item.1}) and~(\ref{cor.item.3}) in Corollary~\ref{cor} are satisfied by 
$f(Z, u) = \beta_0^\intercal k_0(Z_Y, Z_Z, u)$, where $Z = \left(Z_Y^\intercal, Z_Z^\intercal\right)^\intercal$. 
Let $f_n(Z, u) = \widehat{\beta}_n^\intercal \widetilde{k}_n(Z_Y, Z_Z, u)$. 
Standard arguments and the given assumptions 
yield that there exists a $w_1 < \infty$ such that 
$\mathrm{sup}_{Z \in K_{m_1}^q} \mathrm{sup}_{u \in K_{m_2}^d} \left| f_n(Z, u) \right| < w_1$ almost surely 
and 
$$
\operatorname*{sup}_{Z_1, Z_2 \in K_{m1}^q : \|Z_1 - Z_2\| < \delta} 
 \left\| f_n(Z_1, u) - f_n(Z_2, u)\right\|_{K_{m_2}^d} < w_1 \delta
$$
for all $\delta > 0$ and all $n \ge 1$ almost surely. Every subsequence in~(\ref{boot.eq.6}) has a further subsequence 
$n^{\prime \prime}$ on which the convergence in probability to zero can be replaced with almost sure convergence. 
Thus,~(\ref{cor.item.2}) and~(\ref{cor.item.4}) of Corollary~\ref{cor} apply, using the fact that 
minimizing is equivalent to maximizing after a change in sign. Setting 
$\widehat{U}^*_n(Z_Y, Z_Z) = \mathrm{arg \, min}_{u \in K_{m_2}^d} \widehat{\beta}_n^\intercal 
\widetilde{k}_n\left\{ \widetilde{Z}_Y\left(Z_Y\right), \widetilde{Z}_Z\left(Z_Z\right), u\right\}$, 
Corollary~\ref{cor} now yields that 
$$
\operatorname*{sup}_{\left(Z_Y^\intercal, Z_Z^\intercal\right)^\intercal \in K_{m_1}^q} 
 \left\| \widehat{U}^*_{n^{\prime \prime}} (Z_Y, Z_Z) - U_0(Z_Y, Z_Z) \right\| \rightarrow 0
$$
almost surely. Since this is true for every subsequence, we have that 
$$
\operatorname*{sup}_{\left(Z_Y^\intercal, Z_Z^\intercal\right)^\intercal \in K_{m_1}^q} 
 \left\| \widehat{U}^*_n (Z_Y, Z_Z) - U_0(Z_Y, Z_Z) \right\| \xrightarrow[]{P} 0
$$
as $n \rightarrow \infty$. Note that, on $K_2$, $\widehat{U}^*_n(Z_Y, Z_Z) = \widehat{U}_n(Z_Y, Z_Z)$ 
for all $\left(Z_Y^\intercal, Z_Z^\intercal\right)^\intercal \in K_{m_1}^q$. Hence, 
\begin{eqnarray*}
&  & \operatorname*{lim \, sup}_{n \rightarrow \infty} \mathrm{Pr} 
\left\{ \operatorname*{sup}_{\left(Z_Y^\intercal, Z_Z^\intercal\right)^\intercal \in K_{m_1}^q} 
\left\| \widehat{U}_n (Z_Y, Z_Z) - U_0(Z_Y, Z_Z) \right\| > \epsilon \right\} \\
 &  & \hspace{0.2in} \le \operatorname*{lim \, sup}_{n \rightarrow \infty} \Bigg[ \mathrm{Pr}
\left\{ \widehat{U}_n(Z_Y, Z_Z) \in K_2, \operatorname*{sup}_{\left(Z_Y^\intercal, Z_Z^\intercal\right)^\intercal \in K_{m_1}^q} 
\left\| \widehat{U}^*_n (Z_Y, Z_Z) - U_0(Z_Y, Z_Z) \right\| \ge \epsilon \right\} \\ 
 &  & \hspace{0.35in} + \mathrm{Pr}\left\{ \widehat{U}_n(Z_Y, Z_Z) \in K_2^c\right\} \Bigg] \\
 &  & \hspace{0.2in} \le \epsilon.
\end{eqnarray*}
Since $\epsilon$ was arbitrary, we obtain that
$$
\operatorname*{sup}_{\left(Z_Y^\intercal, Z_Z^\intercal\right)^\intercal \in K_{m_1}^q} 
 \left\| \widehat{U}_n (Z_Y, Z_Z) - U_0(Z_Y, Z_Z) \right\| = o_P(1).
$$
Let $B_L(\mathbb{B})$ be the space of all Lipschitz continuous functions mapping $\mathbb{B} \rightarrow \mathbb{R}$ 
which are bounded by 1 and which have Lipschitz constant 1. Let $\mathbb{E}_Z$ be expectation 
with respect to $Z_0^* = \left({Z_Y^*}^\intercal, {Z_Z^*}^\intercal, {Z_A^*}^\intercal\right)^\intercal \sim N(0, \Sigma_0)$. 
Let $B_0\left(Z_0^*\right) = I_0^{-1} \left[ Z_A^* - k_0\left\{ Z_Y^*, Z_Z^*, \widetilde{U}_n\left(Z_0^*\right)\right\}\right]$ and let 
$f \in B_L\left(\mathbb{R}^{d + p}\right)$. Then, 
\begin{eqnarray*}
&  & \left| \mathbb{E}_Z \left[ f\left\{\widetilde{U}_n(Z_0^*), \widetilde{B}_n(Z_0^*)\right\} 
 - f\left\{U_0(Z_0^*), B_0(Z_0^*)\right\} \right] \right| \\
 &  & \hspace{0.2in} \le \left| \mathbb{E}_Z \left[ f\left\{\widetilde{U}_n(Z_0^*), \widetilde{B}_n(Z_0^*)\right\} 
 - f\left\{U_0(Z_0^*), \widetilde{B}_n(Z_0^*)\right\} \right] \right| \\
 &  & \hspace{0.35in} + \left| \mathbb{E}_Z \left[ f\left\{U_0(Z_0^*), \widetilde{B}_n(Z_0^*)\right\} 
 - f\left\{U_0(Z_0^*), B_0(Z_0^*)\right\} \right] \right| \\
 &  & \hspace{0.2in} = \left| \mathbb{E}_Z \left[ f_1\left\{\widetilde{U}_n(Z_0^*)\right\} - f_1\left\{U_0(Z_0^*)\right\}\right] \right| 
 + \left| \mathbb{E}_Z \left[ f_2\left\{\widetilde{B}_n(Z_0^*)\right\} - f_2\left\{B_0(Z_0^*)\right\}\right] \right|
\end{eqnarray*}
for some other $f_1 \in B_L\left(\mathbb{R}^d\right)$ and $f_2 \in B_L\left(\mathbb{R}^p\right)$. Hence, 
\begin{eqnarray*}
 &  & \operatorname*{sup}_{f \in B_L(\mathbb{R}^{d + p})} 
 \left| \mathbb{E}_Z f\left\{ \widetilde{U}_n(Z_0^*), \widetilde{B}_n(Z_0^*)\right\}
 - \mathbb{E}_Z f\left\{ U_0(Z_0^*), B_0(Z_0^*)\right\} \right| \\
 &  & \hspace{0.2in} \le \operatorname*{sup}_{f \in B_L(\mathbb{R}^d)} \left| \mathbb{E}_Z f\left\{\widetilde{U}_n(Z_0^*)\right\} 
 - \mathbb{E}_Z f\left\{U_0(Z_0^*)\right\} \right| \\
 &  &  \hspace{0.35in} + \operatorname*{sup}_{f \in B_L(\mathbb{R}^p)} \left| \mathbb{E}_Z f\left\{\widetilde{B}_n(Z_0^*)\right\} 
 - \mathbb{E}_Z f\left\{B_0(Z_0^*)\right\} \right| \\ 
 &  & \hspace{0.2in} = A_n + B_n, 
\end{eqnarray*}
where we define both $A_n = \mathrm{sup}_{f \in B_L\left(\mathbb{R}^d\right)} \left| \mathbb{E}_Z f\left\{\widetilde{U}_n(Z_0^*)\right\} 
 - \mathbb{E}_Z f\left\{U_0(Z_0^*)\right\} \right|$ and 
$$B_n = \mathrm{sup}_{f \in B_L(\mathbb{R}^p)} \left| \mathbb{E}_Z f\left\{\widetilde{B}_n(Z_0^*)\right\} 
 - \mathbb{E}_Z f\left\{B_0(Z_0^*)\right\} \right|.$$ 
Fix some compact $K_1 \subset \mathbb{R}^q$ such that 
$\mathrm{Pr}\left\{ \left({Z_Y^*}^\intercal, {Z_Z^*}^\intercal\right)^\intercal \in K_1\right\} \ge 1 - \epsilon$. 
Then, 
\begin{eqnarray*}
 &  & \operatorname*{sup}_{f \in B_L(\mathbb{R}^d)} \left| \mathbb{E}_Z\left[ f\left\{ \widetilde{U}_n(Z_0^*)\right\} 
 - \mathbb{E}_Z f\left\{ U_0(Z_0^*)\right\} \right] \right| \\
 &  & \hspace{0.2in} \le \operatorname*{sup}_{f \in B_L(\mathbb{R}^d)} 
 \left| \mathbb{E}_Z 1\left(Z_0^* \in K_1\right) f\left\{\widetilde{U}_n(Z_0^*)\right\} - \mathbb{E}_Z 1\left(Z_0^* \in K_1\right) 
 f\left\{ U_0\left(Z_0^*\right)\right\} \right| \\
 &  & \hspace{0.35in} + 2\mathbb{E}_Z 1\left(Z_0^* \in K_1\right) \\
 &  & \hspace{0.2in} = o_P(1) + 2\epsilon, 
\end{eqnarray*}
which implies that $A_n = o_P(1)$ since $\epsilon$ was arbitrary. For $K_2 \subset \mathbb{R}^{q + p}$ such that 
$\mathrm{Pr}\left(Z_0^* \in K_2\right) \ge 1 - \epsilon$, previous arguments yield that 
$$
\mathrm{sup}_{Z_0^* \in K_2} \left\| \widetilde{B}_n(Z_0^*) - B_0(Z_0^*) \right\| = o_P(1).
$$
As before, we can argue that $B_n = o_P(1) + 2\epsilon$, which implies that $B_n = o_P(1)$ since $\epsilon$ was arbitrary. 
The result follows. 
\end{proof}

\begin{theorem} \label{supp.thm}
Let $H$ be a compact set in a metric space with metric $d$ and let $\mathcal{F}$ be a compact 
subset of $C[H]$ with respect to the uniform norm, $\|\cdot\|_H$. For each $f \in \mathcal{F}$, let 
$u(f) = \mathrm{arg \, max}_{h \in H} f(h)$, where, when the arg max is not unique, we select one element 
of the arg max set either randomly or algorithmically. Suppose also that there exists a closed 
$\mathcal{F}_1 \subset \mathcal{F}$ for which each $f \in \mathcal{F}_1$ has a unique maximum. Then, 
$$
\lim_{\delta \downarrow 0} \, \sup_{f \in \mathcal{F}_1} \, \sup_{g \in \mathcal{F} : \|f - g\|_H < \delta} 
d\left\{u(f), u(g)\right\} = 0.
$$
\end{theorem}

\begin{proof}[Proof of Theorem~\ref{supp.thm}]
Fix $\epsilon > 0$. For each $f \in \mathcal{F}_1$, there exists $\delta_f > 0$ such that 
$$
\sup_{h \in H \cap B_\epsilon\left\{u(f)\right\}^c} f(h) < f\left\{u(f)\right\} - 2 \delta_f,
$$
where $B_\epsilon(u)$ is the open $d$-ball of radius $\epsilon$ around $u$. This follows since the 
compactness of $\mathcal{F}$ ensures that all $f \in \mathcal{F}$ are continuous. Let $g \in \mathcal{F}$ 
be such that $\|f - g\|_H < \delta_f$. Then, 
$f\left\{u(g)\right\} > g\left\{u(g)\right\} - \delta_f \ge g\left\{u(f)\right\} - \delta_f
 > f\left\{u(f)\right\} - 2\delta_f$, 
which implies that $d\left\{u(g), u(f)\right\} < \epsilon$. We have that $\cup_{f \in \mathcal{F}_1} 
\left\{ g \in \mathcal{F} : \|g - f\|_H < \delta_f \right\}$ is an open cover of $\mathcal{F}_1$. 
Since $\mathcal{F}_1$ is compact, there exists a set $\mathcal{F}_1^\epsilon$ such that 
$\mathcal{F}_1^\epsilon$ is finite and 
$$\cup_{f \in \mathcal{F}_1^\epsilon} \left\{ g \in \mathcal{F} : \|g - f\|_H < \delta_f \right\}$$
still covers $\mathcal{F}_1$. Let $\{f_n\} \in \mathcal{F}_1$ and $\{g_n\} \in \mathcal{F}$ be 
sequences. By compactness, every subsequence has a further subsequence $n^{\prime \prime}$ such that 
$f_{n^{\prime \prime}} \rightarrow f_0 \in \mathcal{F}_1$ and 
$g_{n^{\prime \prime}} \rightarrow g_0 \in \mathcal{F}$ so that both $f_0$ and $g_0$ are in a set of the form 
$\left\{ g \in \mathcal{F} : \|g - f\|_H < \delta_f\right\}$ for some $f \in \mathcal{F}_1^\epsilon$. This implies 
that $d\left\{ u(g_0), u(f_0)\right\} < \epsilon$. Since the subsequence was arbitrary, we have that 
$\mathrm{lim \, sup}_{n \rightarrow \infty} d\left\{ u(g_n), u(f_n) \right\} \le \epsilon$. Since $\epsilon$ 
was arbitrary, we now have that $\mathrm{lim \, sup}_{n \rightarrow \infty} d\left\{ u(g_n), u(f_n) \right\} = 0$, 
which proves the result. 
\end{proof}

\begin{corollary} \label{cor}
For $m_1 < \infty$, let $K_{m_1}^q = \left\{ z \in \mathbb{R}^q : \|z\| \le m_1\right\}$. 
Let $(z, u) \mapsto f(z, u)$ and $(z, u) \mapsto f_n(z, u)$ be a fixed function and a sequence of functions, 
respectively, from $K_{m_1}^q \times \mathbb{R}^d$ to $\mathbb{R}$. Suppose there exists $m_2 < \infty$ 
such that for each $z \in K_{m_1}^q$, $u(z) = \mathrm{arg \, max}_{u \in \mathbb{R}^d} f(z, u) < m_2$ 
and is uniquely defined. Suppose also that $u_n(z) = \mathrm{arg \, max}_{u \in \mathbb{R}^d} f_n(z, u) < m_2$ 
for all $n$ large enough, where we allow the arg max to be non-unique, but we randomly or algorithmically select 
one element from the arg max set. Define $K_{m_2}^d$ similarly to $K_{m_1}^q$ and assume that 
\begin{enumerate}
\item $\operatorname*{sup}_{z \in K_{m_1}^q} \operatorname*{sup}_{u \in K_{m_2}^d} \left| f(z, u) \right| < \infty$ \label{cor.item.1}
\item $\operatorname*{lim \, sup}_{n \rightarrow \infty} \operatorname*{sup}_{z \in K_{m_1}^q} \sup_{u \in K_{m_2}^d} 
\left| f_n(z ,u) \right| < \infty$ \label{cor.item.2}
\item $\operatorname*{lim}_{\delta \downarrow 0} \operatorname*{sup}_{z_1, z_2 \in K_{m_1}^q : \|z_1 - z_2\| < \delta} 
\left\| f(z_1, u) - f(z_2, u) \right\|_{K_{m_2}^d} = 0$ \label{cor.item.3} 
\item $\operatorname*{lim}_{\delta \downarrow 0} \operatorname*{sup}_{z_1, z_2 \in K_{m_1}^q : \|z_1 - z_2\| < \delta} 
\left\| f_n(z_1, u) - f_n(z_2, u) \right\|_{K_{m_2}^d} = 0$ \label{cor.item.4} 
\end{enumerate}
for all $n$ large enough. Then, provided $\sup_{z \in K_{m_1}^q} \left\|f_n(z, \cdot) - f(z, \cdot)\right\|_{K_{m_2}^d} \rightarrow 0$, 
$$
\sup_{z \in K_{m_1}^q} \left\| u_n(z) - u(z)\right\| \rightarrow 0
$$ 
as $n \rightarrow \infty$. 
\end{corollary}

\begin{proof}[Corollary~\ref{cor}]
By the Arzel\`a--Ascoli Theorem, there exists a compact $K \subset C[H]$ for $H = K_{m_2}^d$, such that 
both $f(z, \cdot) \in K$ and $f_n(z, \cdot) \in K$ for all $n$ large enough. 
If we let $\mathcal{F}_1 = \left\{f(z, \cdot) : z \in K_{m_1}^q\right\}$, we can directly apply Theorem~\ref{supp.thm} 
to obtain that 
$$
\lim_{\delta \downarrow 0} \, \sup_{z \in K_{m_1}^q} \, \sup_{g \in K : \|g - f(z, \cdot)\|_H < \delta} 
\left\|u(g) - u\left\{f(z, \cdot)\right\} \right\|_H = 0.
$$
Because $\sup_{z \in K_{m_1}^q} \left\| f_n(z, \cdot) - f(z, \cdot)\right\|_{K_{m_2}^d} < \delta$ for all 
$n$ large enough, the result follows. 
\end{proof}

\section*{Appendix B: Three or More Outcomes}

Assume the available data consist of $(\bX_i, A_i, Y_{1,i}, \ldots, Y_{K,i})$, $i= 1, \ldots, n$,
which comprise $n$ independent and identically distributed copies of $(\bX, A, Y_1, \ldots, Y_K)$, 
where $\bX$ and $A$ are as defined previously, and $Y_1, \ldots, Y_K$ are outcomes, with each outcome 
coded so that higher values are better. Assume there exists an unknown utility function 
$U = u(Y_1, \ldots, Y_K)$, where $u : \mathbb{R}^K \rightarrow \mathbb{R}$, such that $u(y_1, \ldots, y_K)$ 
quantifies the ``goodness" of the outcome vector $(y_1, \ldots, y_K)$. As before, let $U^*(d)$ be the 
potential utility under a treatment regime $d$. Let $d_U^\mathrm{opt}$ be the optimal regime for the 
utility defined by $u$, i.e., $\mathbb{E} U^*\left(d_U^\mathrm{opt}\right) \ge \mathbb{E} U^*(d)$ for 
any regime $d$. The goal is to estimate the utility function and the associated optimal regime in 
the presence of more than two outcomes. 

To begin, we assume that the utility function is constant across patients and takes the form 
$u(y_1, \ldots, y_K ; \omega) = \sum_{k = 1}^{K - 1} \omega_k y_k + \left(1 - \sum_{k = 1}^{K - 1} \omega_k\right) y_K$, 
where $\omega = (\omega_1, \ldots, \omega_{K - 1})$ is a vector of parameters with 
$\sum_{k = 1}^{K - 1} \omega_k \le 1$ and $\omega_k \ge 0$ for $k = 1, \ldots, K - 1$. 
Thus, we assume that the utility function is a convex combination 
of the set of outcomes. Let $d^\mathrm{opt}_{\omega}$ be the optimal regime for the utility 
defined by $\omega$. Assume that there exists a true utility function defined by some 
$\omega_0 = (\omega_{1, 0}, \ldots, \omega_{K - 1, 0})$ such that observed decisions were made with 
the intent to maximize $U = u(y_1, \ldots, y_K ; \omega_0)$. Further assume that treatment decisions 
in the observed data follow
$
\mathrm{Pr}\left\{A = d^\mathrm{opt}_{\omega_0}(\bx) | \bX = \bx\right\} 
 = \mathrm{expit} \left(\bx^\intercal \beta \right),
$
where $\beta$ is an unknown parameter. 

Define $Q_{Y_k}(\bx, a) = \mathbb{E}\left(Y_k | \bX = \bx, A = a\right)$, for $k = 1, \ldots, K$. 
Define also
 $$Q_\omega(\bx, a) = \mathbb{E}\left\{u\left(Y_1, \ldots, Y_K; \omega\right) | \bX = \bx, A = a\right\}$$ 
and note that $Q_\omega(\bx, a) = \sum_{k = 1}^{K - 1} \omega_k Q_{Y_k}(\bx, a) 
+ \left(1 - \sum_{k = 1}^{K - 1} \omega_k\right) Q_{Y_K}(\bx, a)$. The Q-functions for each outcome can be estimated 
from the observed data using regression models. Let $\widehat{Q}_{Y_k, n}(\bx, a)$ denote an estimator for $Q_{Y_k}(\bx, a)$. 
Then, an estimator for $Q_\omega(\bx, a)$ is 
$\widehat{Q}_{\omega, n}(\bx, a) = \sum_{k = 1}^{K - 1} \omega_k \widehat{Q}_{Y_k, n}(\bx, a) 
+ \left(1 - \sum_{k = 1}^{K - 1} \omega_k\right) \widehat{Q}_{Y_K, n}(\bx, a)$. 
For any fixed $\omega$, we can compute an estimator for $d_\omega^\mathrm{opt}$ as 
$\widehat{d}_{\omega, n}(\bx) = \operatorname*{arg \, max}_{a \in \mathcal{A}} \widehat{Q}_{\omega, n}(\bx, a)$. 
The pseudo-likelihood is 
\begin{equation} \label{supp-pseudo-likelihood}
\widehat{\mathcal{L}}_n(\omega, \beta) \propto \prod_{i = 1}^n 
 \frac{\mathrm{exp}\left[\bX_i^\intercal \beta 
 1\left\{A_i = \widehat{d}_{\omega, n}(\bX_i)\right\}\right]}
 {1 + \mathrm{exp}\left(\bX_i^\intercal \beta \right)},  
\end{equation}
for a vector $\beta$ and a vector $\omega = (\omega_1, \ldots, \omega_{K - 1})$. For $K = 2$, this 
reduces to the formulation in Section~\ref{comp.single}. Estimators for $\beta$ and $\omega_1, \ldots, \omega_{K - 1}$ 
can be obtained by maximizing the pseudo-likelihood in~(\ref{supp-pseudo-likelihood}). 
Letting $\widehat{\omega}_n = (\widehat{\omega}_{1, n}, \ldots, \widehat{\omega}_{K - 1, n}$ denote the 
maximum pseudo-likelihood estimator for $\omega$, an estimator for the optimal regime is 
$\widehat{d}_{\widehat{\omega}_n, n} (\bx) = \mathrm{arg \, max}_{a \in \mathcal{A}} 
 \widehat{Q}_{\widehat{\omega}_n, n}(\bx, a)$. 

When the number of outcomes is large, maximizing~(\ref{pseudo-likelihood}) using the 
grid search proposed in Section~\ref{fixed.util} is infeasible. However, we can use the Metropolis algorithm 
similar to that proposed for a patient-specific utility function. A patient-specific utility 
function can be accommodated by setting 
$u(y_1, \ldots, y_K ; \bx, \theta) = \sum_{k = 1}^{K - 1}\mathrm{expit}(\bx^\T \theta_k) y_1 
+ \left\{1 - \sum_{k = 1}^{K - 1}\mathrm{expit}(\bx^\T \theta_k)\right\} y_K$ for a vector 
$\theta = (\theta_1^\T, \ldots, \theta_{K - 1}^\T)^\T$ and maximizing the pseudo-likelihood 
using the Metropolis algorithm. 

To examine the performance of the proposed method in the presence of more than two outcomes, 
we use the following generative model. As before, let $\bX = (X_1, \ldots, X_5)^\T$ be 
independent normal random variables with mean 0 and standard deviation 0.5. Let $\epsilon_1$, 
$\epsilon_2$, and $\epsilon_3$ be independent normal random variables with mean 0 and standard 
deviation 0.5. Given treatment assignment, outcomes are generated according to 
$Y_1 = A(4 X_1 - 2 X_2 + 2) + \epsilon_1$, $Y_2 = A(2 X_1 - 4 X_2 - 2) + \epsilon_2$, 
and $Y_3 = 1 + A(X_1 + X_2 + 1) \epsilon_3$. For a fixed $\omega = (\omega_1, \omega_2)$ and 
fixed $\rho \in [0, 1]$, treatment assignment is made according to 
$\mathrm{Pr}\left\{A = d_\omega^\mathrm{opt}(\bx) | \bX = \bx\right\} = \rho$. 

We set $\omega_1 = 0.2$, $\omega_2 = 0.4$, and $\rho = 0.6, 0.8$. 
Table~\ref{est_supp_table1} contains parameter estimates averaged across 
500 replications along with standard deviations (in parentheses) across 
replications. The error rate is the proportion of samples in a testing set that 
were assigned the optimal treatment by the estimated policy. 
\begin{table}[h!]
\centering
\begin{tabular}{cc|cccc}
  \hline
$n$ & $\rho$ & $\widehat{\omega}_{1, n}$ & $\widehat{\omega}_{2, n}$ & $\widehat{\rho}_n$ & Error rate \\ 
  \hline
100 & 0.60 & 0.21 (0.16) & 0.34 (0.20) & 0.63 (0.07) & 0.15 (0.11) \\ 
   & 0.80 & 0.21 (0.07) & 0.42 (0.09) & 0.81 (0.04) & 0.04 (0.03) \\ 
  200 & 0.60 & 0.21 (0.13) & 0.40 (0.17) & 0.62 (0.04) & 0.11 (0.09) \\ 
   & 0.80 & 0.21 (0.04) & 0.41 (0.06) & 0.80 (0.03) & 0.03 (0.02) \\ 
  300 & 0.60 & 0.21 (0.12) & 0.39 (0.16) & 0.62 (0.03) & 0.09 (0.08) \\ 
   & 0.80 & 0.20 (0.03) & 0.41 (0.04) & 0.80 (0.02) & 0.02 (0.01) \\ 
  500 & 0.60 & 0.21 (0.09) & 0.41 (0.12) & 0.61 (0.02) & 0.06 (0.05) \\ 
   & 0.80 & 0.20 (0.02) & 0.40 (0.03) & 0.80 (0.02) & 0.01 (0.01) \\ 
   \hline
\end{tabular}
\caption{Estimation results for simulations where utility and probability of optimal treatment are fixed, with three outcomes.} 
\label{est_supp_table1}
\end{table}
Table~\ref{val_supp_table1} contains estimated values (calculated by generating an 
independent testing set following the same generative model but with decisions made according 
to each policy) of the optimal policy, a policy where the utility function is estimated 
(the proposed method), policies estimated to maximize each outcome individually, and 
standard of care. 
\begin{table}[h!]
\centering
\resizebox{\columnwidth}{!}{
\begin{tabular}{cc|cccccc}
  \hline
$n$ & $\rho$ & Optimal & Estimated utility & $Y_1$ only & $Y_2$ only & $Y_3$ only & Standard of care \\ 
  \hline
100 & 0.60 & 1.38 (0.04) & 1.28 (0.15) & 1.09 (0.06) & 1.00 (0.06) & 0.62 (0.10) & 0.59 (0.14) \\ 
   & 0.80 & 1.39 (0.04) & 1.37 (0.06) & 1.09 (0.06) & 1.00 (0.06) & 0.62 (0.11) & 0.99 (0.13) \\ 
  200 & 0.60 & 1.38 (0.04) & 1.32 (0.12) & 1.09 (0.06) & 1.00 (0.06) & 0.62 (0.08) & 0.60 (0.10) \\ 
   & 0.80 & 1.39 (0.04) & 1.38 (0.05) & 1.09 (0.05) & 1.00 (0.06) & 0.62 (0.09) & 0.98 (0.09) \\ 
  300 & 0.60 & 1.38 (0.04) & 1.34 (0.10) & 1.09 (0.06) & 1.00 (0.06) & 0.63 (0.08) & 0.60 (0.08) \\ 
   & 0.80 & 1.38 (0.04) & 1.39 (0.05) & 1.10 (0.06) & 1.00 (0.06) & 0.63 (0.08) & 0.99 (0.07) \\ 
  500 & 0.60 & 1.39 (0.04) & 1.36 (0.07) & 1.09 (0.05) & 1.00 (0.06) & 0.62 (0.07) & 0.60 (0.06) \\ 
   & 0.80 & 1.39 (0.04) & 1.38 (0.05) & 1.10 (0.06) & 1.00 (0.06) & 0.63 (0.07) & 0.99 (0.06) \\ 
   \hline
\end{tabular}
}
\caption{Value results for simulations where utility and probability of optimal treatment are fixed, with three outcomes.} 
\label{val_supp_table1}
\end{table}
The proposed method results in values close to the true optimal in large samples and 
larger than maximizing each individual outcome across sample sizes. 

\section*{Appendix C: Misspecified Model for the Utility Function}

In this section, we demonstrate that the proposed method achieves reasonable 
performance even in the presence of a misspecified model for the utility function. 
Let $\bX$, $Y$, and $Z$ be generated as above. Let the true underlying utility function be 
$u(y, z; \bx, \theta) = \omega(\bx; \theta) y + \left\{1 - \omega(\bx; \theta)\right\} z$, 
where $\omega(\bx; \theta) = \mathrm{expit}\left(1 + x_1^2 + \bx^\intercal \theta_0\right)$ 
with $\theta_0 = (-0.5, 0, 0, 1, 0.5)^\intercal$. The misspecified model fit to estimate the utility 
function contained only an intercept, $X_1$, $X_2$, $X_3$, and $X_4$. Therefore, these simulations 
represent the setting where one important covariate and a squared term for one covariate 
are omitted from the model for the utility function. Treatment was assigned according to 
$\mathrm{Pr}\left\{A = d_\omega^\mathrm{opt}(\bx) | \bX = \bx\right\} = \mathrm{expit}\left(0.5 + x_1\right)$. 
Table~\ref{val_supp_table2b} contains the estimated value when 
the model for the utility function is correctly specified and when the model 
is incorrectly specified, along with the value of the true optimal policy and 
the standard of care. 
\begin{table}[h!]
\centering
\begin{tabular}{c|cccc}
  \hline
$n$ & Optimal & Correct & Misspecified & Standard of Care \\ 
  \hline
100 & 1.86 (0.07) & 1.61 (0.21) & 1.64 (0.20) & 0.59 (0.23) \\ 
  200 & 1.85 (0.07) & 1.68 (0.16) & 1.69 (0.17) & 0.57 (0.16) \\ 
  300 & 1.86 (0.07) & 1.72 (0.13) & 1.74 (0.13) & 0.57 (0.13) \\ 
  500 & 1.86 (0.07) & 1.77 (0.10) & 1.76 (0.11) & 0.58 (0.10) \\ 
   \hline
\end{tabular}
\caption{First simulation results with a misspecified model for the utility function.} 
\label{val_supp_table2b}
\end{table}
The proposed method produces comparable results regardless of whether the 
utility function is misspecified or not. 

Table~\ref{val_supp_table2a} 
contains results for the same model misspecification, but where the true 
utility function is 
$\omega(\bx; \theta) = \mathrm{expit}\left(1 + 4 x_1^2 + \bx^\intercal \theta_0\right)$ 
with $\theta_0 = (-0.5, 0, 0, 1, 4)^\intercal$, i.e., the coefficients 
for the components that are left out of the misspecified model are larger. 
\begin{table}[h!]
\centering
\begin{tabular}{c|cccc}
  \hline
$n$ & Optimal & Correct & Misspecified & Standard of Care \\ 
  \hline
100 & 2.11 (0.08) & 1.63 (0.28) & 1.64 (0.23) & 0.69 (0.26) \\ 
  200 & 2.11 (0.08) & 1.76 (0.22) & 1.68 (0.18) & 0.67 (0.18) \\ 
  300 & 2.10 (0.07) & 1.84 (0.19) & 1.70 (0.16) & 0.68 (0.15) \\ 
  500 & 2.10 (0.08) & 1.88 (0.16) & 1.73 (0.15) & 0.67 (0.12) \\ 
   \hline
\end{tabular}
\caption{Second simulation results with a misspecified model for the utility function.} 
\label{val_supp_table2a}
\end{table}
When the coefficients of the components left out of the utility function model 
are larger, the proposed method produces better results when the model is 
correctly specified. However, even in the presence of model misspecification, 
the proposed method produces results that improve upon the standard of care. 

\vskip 0.2in
\bibliography{comp}

\end{document}